\documentclass[review]{elsarticle}

\usepackage{lineno,hyperref}
\usepackage{geometry}
\usepackage{amsmath}
\usepackage{tabularx}
\usepackage{booktabs}
\usepackage{placeins}
\usepackage{adjustbox}
\usepackage{graphicx}
\usepackage{pdflscape}
\usepackage{natbib}
\usepackage{amssymb}
\usepackage{tabu}
\usepackage{multirow}
\usepackage{algorithm}
\usepackage{algorithmic}
\usepackage{rotating}
\usepackage{pdflscape}
\usepackage{epstopdf}
\usepackage{color}
\usepackage{url}
\usepackage{supertabular}
\usepackage{blindtext}
\usepackage{multicol}
\usepackage{xcolor}
\usepackage{ltablex}
\usepackage{longtable}
\usepackage{subcaption}
\usepackage{siunitx}

\usepackage[labelfont=bf,justification=raggedright,singlelinecheck=false]
{caption}
\journal{Journal of \LaTeX\ Templates}

\makeatletter
\def\ps@pprintTitle{%
 \let\@oddhead\@empty
 \let\@evenhead\@empty
 \def\@oddfoot{\centerline{\thepage}}%
 \let\@evenfoot\@oddfoot}
\makeatother

\begin{document}


\begin{frontmatter}


\title{Chaos embedded opposition based learning for gravitational search algorithm}

\author{Susheel Kumar Joshi}
\ead{sushil4843@gmail.com}


\address{South Asian University, New Delhi}

\begin{abstract}

Due to its robust search mechanism, Gravitational search algorithm (GSA) has achieved lots of popularity from different research communities. However, stagnation reduces its searchability towards global optima for rigid and complex multi-modal problems. This paper proposes a GSA variant that incorporates chaos-embedded opposition based learning into the basic GSA for the stagnation free search. Additionally, a sine-cosine based chaotic gravitational constant is introduced to balance the trade-off between exploration and exploitation capabilities more effectively. The proposed variant is tested over $23$ classical benchmark problems, $15$ test problems of CEC $2015$ test suite, and $15$ test problems of CEC $2014$ test suite. Different graphical, as well as empirical analyses, reveal the superiority of the proposed algorithm over conventional meta-heuristics and most recent GSA variants.

\end{abstract}


\end{frontmatter}


\section{Introduction}\label{sec:intro}
 
Due to having flexible and robust search mechanisms, meta-heuristic optimization algorithms have been gained a substantial amount of recognition in various fields of scientific research, engineering, and technology. Their derivative-free search mechanism makes them superior to the conventional gradient-based optimization methods over the rigid and complex real-world NP-hard optimization problems. The search process of these algorithms solely depends upon their two fundamental abilities named exploration and exploitation. The former investigates the search space to find more sub-optimal regions regarding the given fitness landscape. While the latter works as a local search around the neighborhood of the optimal solutions belong to the most promising regions explored so far. A balanced trade-off between exploration and exploitation provides a robust and efficient search mechanism through which the algorithm obtains the quality solutions. In contrast, the unbalanced trade-off between these two capabilities works against the search requirements. In the middle and terminal phases of the search process, the excessive exploration increases the search intensity of the algorithm through which the search may skip the optimal regions unnecessarily. While excessive exploitation in the initial or the middle phases of the search process causes stagnation in the local optimal regions. To overcome these flaws, the algorithm can be developed by different approaches. Tuning the attached parameters of the algorithm, introducing self-adaptiveness in different entities of the algorithm, hybridize two or more algorithms, introducing more efficient neighborhood structures, and adopting more flexible position update strategies are some of these approaches in this regard. Opposition based learning (OBL) approach is one of the most recognized research fields that can be viewed as a position update strategy for the candidate solutions regarding meta-heuristic algorithms. OBL helps the algorithm to find the optimal regions of the fitness landscape by considering a position and its opposite simultaneously. In literature, several studies have been conducted to hybridize OBL approaches with different meta-heuristics to develop their more robust variants. In this regard, Rahnamayan et. al. \cite{rahnamayan2008opposition} introduced OBL in differential evolution (DE) algorithm to improve its exploitation ability. In \cite{shan2016modified}, Shan et. al. incorporated OBL approach with Bat Algorithm (BA) for a more diversified search. In a similar study, Zhou et. al. \cite{zhou2017opposition} improved the search diversity of memetic algorithm (MA) by using OBL. Sapre et. al. \cite{sapre2019opposition} proposed an efficient novel variant of moth flame optimization (MFO) embedded with OBL. Sarkhel et. al. \cite{sarkhel2017novel} used OBL to develop a novel variant of Harmony Search (HS) having fast convergence speed. In a similar study \cite{verma2016opposition}, OBL is used to improve the performance of Firefly Algorithm (FA). To overcome the possibility of stagnation occurrence, Wang et. al. \cite{wang2011enhancing} proposed an enhanced particle swarm optimization (PSO) algorithm associated with a generalized OBL approach along with a mutation operator. In \cite{ahandani2012opposition}, Ahandani et. al. used OBL to develop a stagnation avoidance mechanism for DE. Feng et. al. \cite{feng2018opposition} improved the performance of monarch butterfly optimization (MBO) algorithm by using a generalized (OBL) approach. In the past few years, OBL approach has gained more attention as a stagnation avoidance tool for meta-heuristics. The recent developments of grasshopper optimization algorithm (GOA) \cite{ewees2018improved}, sine cosine algorithm (SCA) \cite{gupta2019hybrid}, hybrid model of whale optimization algorithm (WOA) and DE \cite{ewees2021new}, elephant herding optimization (EHO) \cite{muthusamy2021improved}, cuckoo optimization algorithm \cite{hamidzadeh2021feature}, DE \cite{choi2021fast}, grey wolf optimizer (GWO)\cite{bansal2021better}, yin-yang pair optimization algorithm \cite{wang2021orthogonal}, moth swarm optimization (MSO) \cite{oliva2021opposition}, laplacian equilibrium optimizer (LEO) \cite{dinkar2021opposition}, salp swarm algorithm (SSA) \cite{tubishat2020improved}, harris hawks optimization (HHO) \cite{gupta2020opposition, sihwail2020improved} indicate the tremendous growth of OBL in meta-heuristics.

Gravitational search algorithm (GSA) \cite{ras} is an efficient and robust meta-heuristic algorithm that follows the rules of gravity. Like other meta-heuristic frameworks, GSA also has its hyper-parameters that heavily affect its performance. Gravitational constant $G$ is the most sensitive entity of the GSA model which effectively controls the trade-off between exploration and exploitation ability of the algorithm. $\alpha$ and $G_0$ are two constant parameters associated with $G$ and therefore affect the perofrmance of the algorithm. As far as the tuning of these mentioned parameters is concerned, several GSA variants have been developed.  Mirjalili et al. \cite{mirjalili2017chaotic} proposed a self-adaptive gravitational constant which can adopt different chaotic nature through different chaotic maps. Bansal et al.\cite{bansal2018fitness} proposed a dynamic gravitational constant 
based GSA variant in which different agents get different gravitational constants according to their current fitnesses. Wang et.al. \cite{wang2019hierarchical} proposed a modified gravitational constant for a hierarchical GSA model to get the maximum knowledge about the population topology. To accelerate the convergence speed and provide a diversified search, several studies  \cite{li2012parameters, gao2014gravitational,mittal2016chaotic, song2017multiple,wang2020gravitational} hybridized the randomness and dynamic nature of different chaotic maps into the stochastic search mechanism of GSA. Joshi et. al. \cite{joshi2020parameter} proposed a method of parameter tuning based on the topological characteristics of the fitness landscape. In this study, the proposed method is used to tune the parameter $\alpha$ in $G$. Pelusi et. al. \cite{pelusi2020improving} tuned the gravitational constant by introducing hyperbolic sine functions in GSA. Sombra et al. \citep{FSalpha} employed a fuzzy approach to make $\alpha$ as a self-adaptive parameter in $G$. In a similar study, Saeidi-Khabisi et al. \cite{Fuzzyalpha} integrated fuzzy logic controller in the GSA model to control $\alpha$ as per the search requirement. Chaoshun Li et al. \cite{MGSA} proposed a dynamic $\alpha$ incorporated with a hyperbolic function to reduce the possibility of premature convergence. Sun et al. \cite{sun2018stability} used the agent's position and its fitness for a self-adaptive $\alpha$ which further improves the performance of the GSA model. In a recent study, Joshi et. al. \cite{joshi2021novel} proposed a self-adaptive fitness-distance-based gravitational constant which scales the agent's next move in each direction of its neighbours. In literature, a few studies have been conducted regarding the use of OBL to enhance the performance of GSA. In the pioneering work, Shaw et. al. \cite{shaw2012novel, shaw2014solution} utilized OBL for generating the initial swarm and for improving the search mechanism through specific random iterations. In a subsequent study, Bhowmik et. al. \cite{bhowmik2015solution} used OBL to improve the performance of GSA search mechanism under the multi-objective framework. In this paper, the hybrid search mechanism of GSA and OBL is considered to develop a more robust GSA variant having the following novelties:

\begin{itemize}

\item The proposed variant introduces a chaotic sequence-based stochastic OBL approach to enhance the searchability of basic GSA for obtaining the most promising regions of the fitness landscape with low computational efforts.  

\item The main merit of the proposed strategy is its least computational expensive nature. For a specific iteration, the considered OBL approach generates the opposite for a randomly selected candidate solution only and leaves the remaining candidate solutions as their original positions. The random selection of a candidate solution for applying OBL provides additional exploration to the embedded GSA model through which it significantly deals with stagnation. On the other hand, considering a single candidate solution out of the whole population preserves the useful information regarding the fitness landscape provided by the remaining
candidate solutions. Due to the same reason, the attached OBL uses a single function evaluation for its implementation.  
\item To overcome the issues regarding the shape of the basic gravitational constant function $G(t)$ in GSA, a sine-cosine based chaotic gravitational constant is proposed for avoiding the local entrapments of the candidate solutions during the middle phase of the search process.
  
\end{itemize}

The remainder of this paper is organized as follows. Section \ref{sec:gsa} briefly describes the frameworks of GSA and OBL. In Section \ref{sec:3}, a detailed introduction of the proposed variant is given. The experimental setting and simulation results are presented in Section \ref{sec:exp}. Finally, Section \ref{sec:conclusion} concludes the paper.



\section{Basic gravitational search algorithm}\label{sec:gsa}
\vspace*{1.5px}
Gravitational Search Algorithm (GSA) is a new meta-heuristic technique for optimization developed by Rashedi et al \cite{ras}. This  algorithm is inspired by the law of gravity and the law of motion.

 The GSA algorithm can be described as follows:\\
Consider the population of $N$ agents (candidate solutions), in which each agent $X_{i}$ in the search space $\mathbb{S}$ is defined as:
\begin{equation}\label{eq1}
  X_{i}=(x_{i}^{1},.....,x_{i}^{d},.....,x_{i}^{n}), ~~ \forall ~i=1,2,.....,N
\end{equation}

Here, $X_{i}$ shows the position of $i^{th}$ agent in $n$-dimensional search space $\mathbb{S}$. The mass of each agent depends upon its fitness value calculated as below:\\
\begin{equation}
q_i(t)=\frac{fit_i(t)-worst(t)}{best(t)-worst(t)}
\end{equation}
\begin{equation}
M_i(t)=\frac{q_i(t)}{\sum_{j=1}^{N}q_j(t)}, ~~\forall ~i=1,2,.....,N
\end{equation}
Here,\\
$fit_i(t)$ is the fitness value of agent $X_i$ at iteration $t$,\\
 $M_i(t)$ is the mass of agent $X_i$ at iteration $t$.\\
 Worst(t) and best(t) are worst and best fitness of the current population, respectively.

The acceleration of $i^{th}$ agent in $d^{th}$ dimension is denoted by $a_i^{d}(t)$ and defined as: \\
\begin{equation}
a_i^{d}(t)=\frac{F_i^{d}(t)}{M_i(t)}
\end{equation}

Where $F_i^{d}(t)$ is the total force acting on the ${i}^{th}$ agent by a set of Kbest heavier masses in ${d}^{th}$ dimension at iteration $t$. $F_i^{d}(t)$ is calculated as:\\
\begin{equation}
F_{i}^{d}(t) = \sum _{j \in K_{best}, j \neq i}rand_{j} \times F_{ij}^{d}(t)
\end{equation}
Here, $K_{best}$ is the set of first $K$ agents with the best fitness values (say $K_{best}$ agents) and biggest masses and $rand_{j}$ is a uniform random number between $0$ and $1$. Cardinality of $K_{best}$, i.e. $K_{best}$ is a linearly decreasing function of time. The value of $K_{best}$ will reduce in each iteration and at the end only one agent will apply force to the other agents. At the ${t}^{th}$ iteration, the force applied on agent $i$ by agent $j$ in the ${d}^{th}$ dimension is defined as:
\begin{equation}\label{eq:force}
F_{ij}^{d}(t)=G(t)\frac{M_i(t)M_j(t)}{R_{ij}+\epsilon } (x_i^{d}(t)-x_j^{d}(t))
\end{equation}
Here, $R_{ij}(t)$ is the Euclidean distance between two agents, $i$ and $j$. $\epsilon$ ($\epsilon > 0$) is a small number.
Finally, the acceleration of an agent in $d^{th}$ dimension is calculated as:
\begin{equation}\label{eq:acc}
a_i^{d}(t)=\sum_{j\in K_{best},j\neq i}rand_{j}G(t)\frac{M_j(t)}{R_{ij}+\epsilon } (x_i^{d}(t)-x_j^{d}(t)),
\end{equation}
$d = 1, 2, . . . , n$ and $i = 1, 2, . . . , N$.\\
$G(t)$ is called gravitational constant and is a decreasing function of time (iteration):
\begin{equation}\label{eq:gc}
G(t)=G_{0}e^{-\alpha \frac{t}{T}}
\end{equation}
$G_{0}$ and $\alpha$ are constants and set to $100$ and $20$, respectively. $T$ is the total number of iterations.

The velocity update equation of an agent $X_i$ in $d^{th}$ dimension is given below:\\
\begin{equation}\label{eq:vel}
v_{i}^{d}(t+1)=rand_i\times{v_{i}^{d}(t)}+a_{i}^{d}(t)
ss\end{equation}
Based on the velocity calculated in equation (\ref{eq:vel}), the position of an agent $X_i$ in $d^{th}$ dimension is updated using position update equation as follow: \\ 
\begin{equation}\label{eq:pos}
x_{i}^{d}(t+1)=x_{i}^{d}(t)+v_{i}^{d}(t+1)
\end{equation}
where $v_{i}^{d}(t)$ and $x_{i}^{d}(t)$ represent the velocity and position of agent $X_i$ in $d^{th}$ dimension,  respectively. $rand_{i}$ is uniformly distributed random number in the interval $\left[0,1\right]$.
 
\subsection{Opposition based learning}\label{OBL}

Meta-heuristics are stochastic optimization algorithms that seek near-optimal solutions for a given specific fitness landscape iteratively. The search process starts from an initial swarm of randomly generated candidate solutions that follow certain search rules to update themselves and finally converge towards the most promising region of the landscape. The main aim of the algorithm is to provide a robust search mechanism through which these candidate solutions get closer to the optimal region. However, the unbalanced trade-off between exploration and exploitation can be responsible for either stagnation or skipping the true optima. In both cases, the required closeness is not achievable. These issues can be resolved by using OBL that consider both position and opposite position of the candidate solution simultaneously. If an individual candidate solution is far away from the unknown optimal region, its opposite position may be very close to that optimal region. In other words, following a direction and its opposite simultaneously can make a diversified and robust search mechanism that provides higher possibilities of getting optimal regions.

In the $n$-dimensional search space, the opposite point (or position) $Y_i(y_i^1,...,y_i^j,...,y_i^n)$ of a candidate solution (reference position) $X_i(x_i^1,...,x_i^j,...,x_i^n)$ is defined as:   

\begin{equation}
y_i^j=lb_j+ub_j-x_i^j,~x_i^j\in [lb_j,ub_j], j=1,2,...n
\end{equation}
Here $lb_j$ and $ub_j$ are the upper and lower bounds of the given search space in its $j^{th}$ dimension respectively. Regarding a specific reference position, OBL can follow different strategies to produce different points having different degrees of opposition. An opposite point with the strongest degree always makes a complete symmetry with its corresponding reference position. While the weakest degree provides flexibility to the opposite point for choosing any position except its reference point. In the context of OBL attached meta-heuristic algorithm, the degree of opposition heavily affects its exploration ability. A proper control between the strongest and weakest degrees provides the most appropriate opposites to the meta-heuristic search mechanism for better exploration. In this regard, stochastic OBL approaches like Quasi OBL \cite{rahnamayan2007quasi} and generalized OBL (GOBL) \cite{wang2011enhancing} play significant roles to control the degree of opposition using randomness guided by uniform distribution. However, uniformly distributed randomness provides an equal chance to select a position which further makes possibilities of either high search intensity or low exploration ability. Due to the high search intensity, the possibility of generating opposite point outside the predefined range is also there. To eliminate these issues, several studies \cite{yang2021opposition, houssein2021enhanced, nasser2021adaptive} have 
used other formats of GOBL with the same uniformly distributed randomness. While some other studies have been used non-uniform distribution like beta distribution \cite{park2015stochastic} and chaotic sequence \cite{singh2021chaotic} to replace uniform distribution in GOBL.

\section{The proposed GSA variant}\label{sec:3}
\subsection{The sine cosine based chaotic gravitational constant}\label{subsection:seq}

\begin{center}
 \begin{figure}
    \includegraphics[width=\linewidth,height=1.5in]{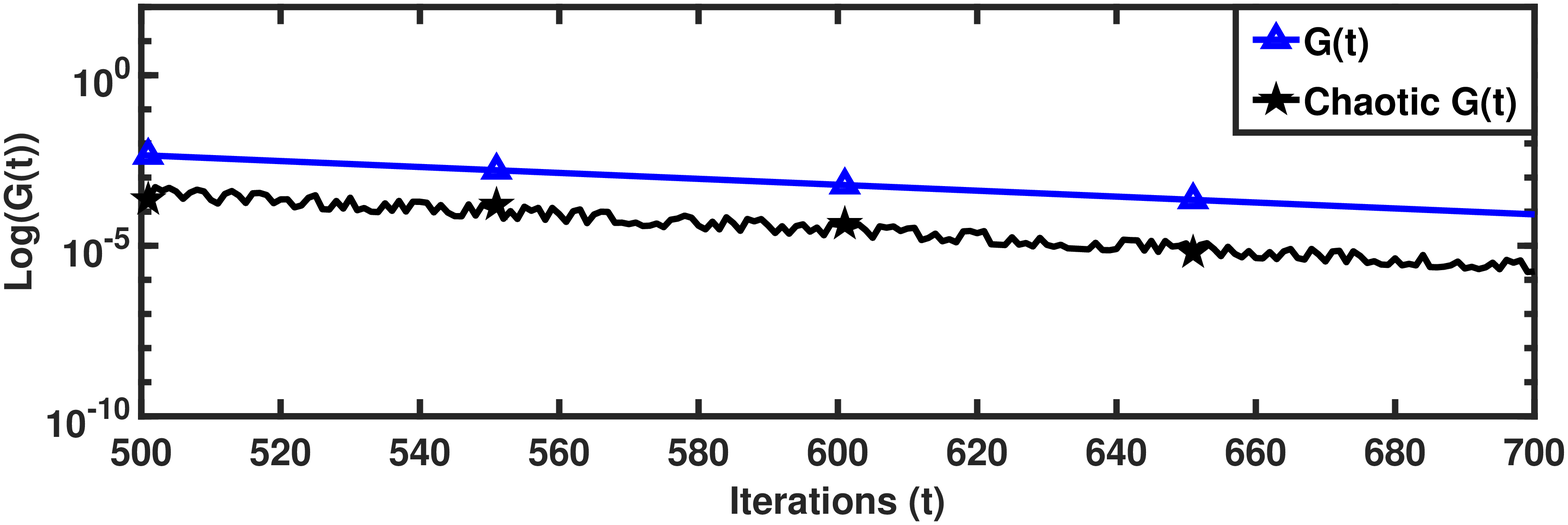}
    \caption{Comparison between the basic $G(t)$ and the introduced chaotic $G(t)$ during the middle phase of the search under unimodal landscape} \label{fig:com_G}
    \end{figure}
\end{center}

Gravitational constant $G(t)$ is one of the foremost entities of the GSA model which is solely responsible for the trade-off between exploration and exploitation. The step size of the candidate solutions follows the shape of $G(t)$, which decreases exponentially as iteration increases. This is a suitable approach through which the candidate solutions get a large step size in the initial phase of the search process for better exploration. As per the decreasing value of $G(t)$, the step size of candidate solutions decreases and the search process changes its mode from exploration to exploitation accordingly. During this shifting, somewhere in the middle phase of the search process, the value of $G(t)$ does not change significantly (refer to Figure \ref{fig:com_G}). Consequently, the step sizes of the candidate solutions do not change as per the search requirements which further causes either stagnation or skipping the true optima. To overcome this issue, a chaotic gravitational constant is introduced by incorporating a sine-cosine based chaotic $\alpha$ into it. Algorithm \ref{algorithm:Chaos_G} describes the pseudo-code to calculate the proposed chaotic $\alpha$. In Algorithm \ref{algorithm:Chaos_G}, $cs$ denotes the chaotic sequence which is generated by the well known logistic map defined as:

\begin{equation}
cs(t+1)=a\times cs(t)(1-cs(t))
\end{equation}     
Here a=$4$ and the initial value of the sequence is $0.7$ ($cs(1)$=$0.7$). $A$, $B$,  and $G_0$ are set to $2$, $25$, and $100$ respectively. While $t$ and $T$ have their usual meaning. Figure \ref{fig:com_G} illustrates the comparative behaviour between the chaotic $G(t)$ associated with chaotic $\alpha$ and the basic $G(t)$ over the unimodal landscape (sphere function). For the middle phase of the search process, the abrupt changes in chaotic $G(t)$ provide different step sizes to the candidate solutions which further avoid the possibilities of stagnation in the local optima.
 
 \begin{algorithm}[tb!]
  \begin{algorithmic}[1]
   \caption{Chaotic $\alpha$ for the proposed chaotic gravitational constant:}
   \label{algorithm:Chaos_G}
   \STATE Initialize $A$, $B$, $T$, $G_0$, $cs$
\STATE Calculate $r=A \times (1-\frac{t}{T})$ 
\IF{($rand<0.5$)}
\STATE $\alpha=B-r\times sin(cs(t))$
 \ELSIF{($rand>0.5$)}
\STATE $\alpha=B+r\times cos(cs(t))$
\ENDIF   
                     
  \end{algorithmic}
 \end{algorithm}

\subsection{Incorporate chaos embedded OBL into GSA}

In this study, the chaotic sequence $cs$ described in the 
Section \ref{subsection:seq} is employed to produce a stochastic OBL defined as:       

\begin{equation}\label{eqn:obl}
y_i^j= cs(t)\times(lb_j+ub_j-x_i^j),~x_i^j\in [lb_j,ub_j], j=1,2,...n
\end{equation}

Here $lb_j$ and $ub_j$ are the lower and upper bound of the $j^{th}$ dimension variable in the current swarm respectively. The considered chaotic sequence makes a better trade-off between the strongest and weakest degree of oppositions which further helps to generate the most appropriate opposite corresponding to a given candidate solution. Since the considered chaotic sequence is bounded therefore the generated opposites always remain inside the search space. It is always a crucial task to combine an OBL strategy with a meta-heuristic to get the optimality in terms of performance as well as computational cost. In most of the OBL attached meta-heuristics, all the candidate solutions of a specific population are considered for calculating their opposites followed by the greedy approach. Changing all the positions to their respective opposites may discard useful information of the fitness landscape provided by certain good-fitted original individuals. While the additional function evaluations make this consideration more computationally expensive. Although elite OBL reduces the computational complexity by calculating the opposites of elite candidate solutions only, there is always a chance to waste the fitness evaluations if suitable opposites can no longer be discovered somewhere middle and the terminal phases of the search process. To address these issues, the proposed chaos embedded OBL (refer to Equation (\ref{eqn:obl})) is applied to a single candidate solution selected from a specific population of the GSA model randomly. This random selection provides additional exploration to the search  which further removes the possibilities of stagnation. On the other hand, considering a single candidate solution needs low computational efforts and preserves useful information about the remaining regions of the fitness landscape. This combination of GSA with the chaotic OBL, called COGSA (chaos embedded opposition based learning for gravitational search algorithm ) is described in Algorithm \ref{algorithm:Chaos_GSA}.

 \begin{algorithm}[tb!]
  \begin{algorithmic}[1]
   \caption{COGSA algorithm:}
   \label{algorithm:Chaos_GSA}
   \STATE Initialize the candidate solutions randomly
   \STATE Initialize their velocities
   \STATE Initialize $T$, $G_0$
   \STATE Generate chaotic sequence $cs$ by logistic map      
    \WHILE{$(t < T)$}    
      \STATE Evaluate fitness of each candidate solution
      \STATE Find best and worst fitness of the current swarm
     \STATE Calculate masses
\STATE Calculate the chaotic $\alpha$ by using Algorithm \ref{algorithm:Chaos_G}
\STATE Calculate $G(t)=G_0\times e^{\frac{-\alpha\times t}{T}}$

\STATE Find acceleration
\STATE Update velocities and positions of agents
\STATE Choose a candidate solution $X_{k}(x_k^1,...,x_k^j,...,x_k^n)$ from the current swarm randomly

\FOR{$(j=0,j<n,j++)$} 
\STATE $y_k^j=cs(t)\times(lb_j+ub_j-x_k^j)$
\ENDFOR
\STATE Select the best fit between $X_k$ and $Y_k$ for the updated swarm  

\ENDWHILE         
                     
  \end{algorithmic}
 \end{algorithm}

\section{Results and Discussion}\label{sec:exp}

\subsection{Testbeds under consideration}
To evaluate the overall performance of the proposed GSA variant, two testbeds (Testbed $1$ and Testbed $2$) having a wide range of benchmarks problems are considered. The first testbed (Testbed $1$) is a well-known set of $23$ benchmark problems ($f_1-f_{23}$) which is further classified into three  categories as per their topological characteristics. The first category contains seven unimodal test problems from $f_1$ to $f_7$ which evaluate the exploitation ability of the algorithm. The second group is formed by six multimodal test problems from $f_8$ to $f_{13}$ which check the stagnation avoidance mechanism of the algorithm. The third group is a set of ten fixed dimensional multimodal test problems ($f_{14}-f_{23}$). Table \ref{table:Testbed1} presents the properties of these test problems like objective function, search space, optimal value, and topological characteristics. In this table, $n$ indicates the dimension of thirteen scalable test problems from $f_1$ to $f_{13}$. The test problems from $f_{14}$ to $f_{23}$ are not scalable due to their fixed dimensions. Testbed $2$, on the other hand, is a set of $30$ rigid and complex test problems of CEC $2015$ \cite{liang2014problem} and CEC $2014$ 
\cite{liang2013problem} test suites. Out of $30$, $15$ problems ($g_1$, $g_2$, $g_5$, $g_6$, $g_7$, $g_{13}$, $g_{14}$, $g_{15}$, $g_{18}$, $g_{19}$, $g_{20}$, $g_{21}$, $g_{22}$, $g_{23}$, and $g_{24}$) are from CEC $2015$ test suite while the remaining $15$ problems ($g_3$, $g_4$, $g_8$, $g_9$, $g_{10}$, $g_{11}$, $g_{12}$, $g_{16}$, $g_{17}$, $g_{25}$, $g_{26}$, $g_{27}$, $g_{28}$, $g_{29}$, and $g_{30}$) are taken from CEC $2014$ test suite. These $30$ test problems are further classified into four groups: unimodal test problems ($g_1$-$g_4$), multi-modal test problems ($g_5$-$g_{12}$), hybrid test problems ($g_{13}$-$g_{17}$), and composite test problems ($g_{18}$-$g_{30}$). Due to having diverse properties like shifted and rotated fitness landscapes, various local optimal points, ill-conditioning, and impurity, these $30$ problems produce the toughest topological challenges to an algorithm for finding the optimal regions. All the problems of Testbed $2$ are considered in the $30$-dimensional search space with a range between $-100$ to $100$.

\begin{center}
\setlength{\LTleft}{-20cm plus -1fill}
\setlength{\LTright}{\LTleft}
\begin{longtable}{p{11.2cm}p{3.2cm}p{2.2cm}p{1.4cm}}
 \caption{Testbed $1$ ( \textbf{S}: Search space,  \textbf{n}: Dimension,  \textbf{$f_{min}$}: Optimal value, \textbf{C}: Characteristic, \textbf{U}: Unimodal, \textbf{M}: Multimodal, \textbf{MFD}: Multimodal with fixed dimension)}
 \label{table:Testbed1}
\endfirsthead
\caption* {\textbf{Table \ref{table:Testbed1} Continued:}}
\endhead
\toprule[0.9pt]
 \textbf{Test problem} &   \textbf{S}  & \textbf{$f_{min}$}  &\textbf{C}  \\
 \midrule[0.9pt]

$f_1(x)=\sum_{i=1}^{n}x_{i}^{2}$  & $[-100,100]^{n}$ & 0 & $U$ \\[1ex]
 
 $f_2(x)=\sum_{i=1}^{n}|x^{2}_{i}|$+ $\prod_{i=1}^{n}|x_{i}|$  & $[-10,10]^{n}$ & 0 & $U$ \\[1ex]
 
 $f_3(x)=\sum_{i=1}^{n}\left(\sum_{j-1}^{i}x_{j}\right)^{2}$  & $[-100,100]^{n}$ & 0 & $U$ \\[1ex]
 
 $f_4\left(x\right)=\max_{i}\left\{ |x_{i}|,1\leq i\leq n\right\} $  & $[-100,100]^{n}$ & 0 & $U$ \\[1ex]
 
 $f_5(x)=\sum_{i=1}^{n-1}\left[100\left(x_{i+1}-x_{i}^{2}\right)^{2}+\left(x_{i}-1\right)^{2}\right]$  & $[-30,30]^{n}$ & 0& $U$\\[1ex]
 
 $f_6(x)=\sum_{i=1}^{n}\left(\left[x_{i}+0.5\right]\right)^{2}$  & $[-100,100]^{n}$ & 0 & $U$ \\[1ex]
 
        \bottomrule[0.9pt]
 \pagebreak
 \toprule[0.9pt]
\textbf{Test problems} &    \textbf{S}  & \textbf{$f_{min}$}  &\textbf{C} \\
 \midrule[0.9pt]   
 
 $f_7(x)=\sum_{i=1}^{n}ix_{i}^{4}+random[0,1)$  & $[-1.28,1.28]^{n}$ & 0 & $U$  \\[1ex]

  $f_8(x)=\sum_{i=1}^{n}-x_{i}\sin\left(\sqrt{|x_{i}|}\right)$  & $[-500,500]^{n}$ & $-$418.9829$\times n$  & $M$ \\ [1ex]

  $f_9(x)$=$\sum_{i=1}^{n}\left[x_{i}^{2}-10\cos\left(2\pi x_{i}\right)+10\right]$  & $[-5.12,5.12]^{n}$ & 0 & $M$ \\[1ex]

  $f_{10}(x)=-20\exp(-0.2\sqrt{\frac{1}{n}\sum_{i=1}^{n}x_{i}^{2}})-\exp\left(\frac{1}{n}\sum_{i=1}^{n}cos\left(2\pi x_{i}\right)\right)+20+e$ & $[-32,32]^{n}$ & 0& $M$ \\ [1ex]
  
  $f_{11}(x)=\frac{1}{4000}\sum_{i=1}^{n}x_{i}^{2}-\prod_{i=1}^{n}\cos\left(\frac{x_{i}}{\sqrt{i}}\right)+1$  & $[-600,600]^{n}$ & 0 & $M$ \\[1ex]

  $f_{12}(x)=\frac{\pi}{n}\left\{ 10\sin\left(\pi y_{1}\right)+\sum_{i=1}^{n-1}(y_{i}-1)^{2}\left[1+10\sin^{2}(\pi y_{i+1})\right]+ 
  (y_{n}-1)^{2}\right\} +\sum_{i=1}^{n}u(x_{i},10,100,4)$\\
  $y_{i}=1+\frac{x_{i}+1}{4}u(x_{i},a,k,m)=\begin{cases}
  k(x_{i}-a)^{m} & x_{i}>a\\
  0-a & <x_{i}<a\\
  k(-x_{i}-a)^{m} & x_{i}<-a
  \end{cases}$  & $[-50,50]^n$ & 0 & $M$\\[1ex]

  $f_{13}(x)=0.1\left\{ \sin^{2}(3\pi x_{1})+\sum_{i=1}^{n}\left(x_{i}-1\right)^{2}\left[1+\sin^{2}(3\pi x_{i}+1)\right]+ \right.\newline\left.
  (x_{n}-1)^{2}\left[1+\sin^{2}(2\pi x_{n})\right]\right\} +\sum_{i=1}^{n}u(x_{i},5,100,4)$   
  &  $[-50,50]^{n}$ & 0 & $M$\\[1ex]

  $f_{14}(x)=\left(\frac{1}{500}+\sum_{j=1}^{25}\frac{1}{j+\sum_{i=1}^{2}\left(x_{i}-a_{ij}\right)^{6}}\right)^{-1}$ &  $[-65,65]^{2}$ & 0.998 & $MFD$ \\[1ex]

    $f_{15}(x)=\sum_{i=1}^{11}\left[a_{i}-\frac{x_{1}\left(b_{i}^{2}+b_{i}x_{2}\right)}{b_{i}^{2}+b_{i}x_{3}+x_{4}}\right]^{2}$ &  $[-5,5]^{4}$ & 0.00030 & $MFD$ \\[1ex]
    
    $f_{16}(x)=4x_{1}^{2}-2.1x_{1}^{4}+\frac{1}{3}x_{1}^{6}+x_{1}x_{2}-4x_{2}^{2}+4x_{2}^{4}$ & $[-5,5]^{2}$ & $-$1.0316 & $MFD$ \\[1ex]

    $f_{17}\left(x\right)=\left(x_{2}-\frac{5.1}{4\pi^{2}}x_{1}^{2}+\frac{5}{\pi}x_{1}-6\right)^{2}+10\left(1-\frac{1}{8\pi}\right)\cos x_{1}+10$ &  $[-5,10]\times [0,15]^{2}$ & 0.398 & $MFD$ \\[1ex]
    
    $f_{18}(x)=\left[1+\left(x_{1}+x_{2}+1\right)^{2}\left(19-14x_{1}+3x_{1}^{2}-14x_{2}+6x_{1}x_{2}+3x_{2}^{2}\right)\right]$\\ $\left[30+\left(2x_{1}-3x_{2}\right)^{2}\left(18-32x_{1}+12x_{1}^{2}+48x_{2}-36x_{1}x_{2}+27x_{2}^{2}\right)\right]$ &  $[-2,2]^{2}$ & 3 & $MFD$\\[1ex]

    $f_{19}(x)=-\sum_{i=1}^{4}c_{i}\exp\left(-\sum_{j=1}^{3}a_{ij}\left(x_{j}-p_{ij}\right)^{2}\right)$ &  $[0,1]^{3}$ & $-$3.86 & $MFD$\\[1ex]
    
    $f_{20}(x)=-\sum_{i=1}^{4}c_{i}\exp\left(-\sum_{j=1}^{6}a_{ij}\left(x_{j}-p_{ij}\right)^{2}\right)$ &  $[0,1]^{6}$ & $-$3.32 & $MFD$ \\[1ex]
    
    $f_{21}(x)=-\sum_{i=1}^{5}\left[(X-a_{i})\left(X-a_{i}\right)^{T}+c_{i}\right]^{-1}$ & $[0,10]^{4}$ & $-$10.1532 & $MFD$  \\[1ex]
    
    $f_{22}(x)=-\sum_{i=1}^{7}\left[(X-a_{i})\left(X-a_{i}\right)^{T}+c_{i}\right]^{-1}$ &  $[0,10]^{4}$ & $-$10.4028 & $MFD$ \\[1ex]
    
    $f_{23}(x)=-\sum_{i=1}^{10}\left[(X-a_{i})\left(X-a_{i}\right)^{T}+c_{i}\right]^{-1}$ &  $[0,10]^{4}$ & $-$10.5363 & $MFD$ \\[1ex]
  
\bottomrule[0.9pt]
\end{longtable}
\end{center}

\subsection{Experimental setting}
In order to investigate the search abilities of the proposed COGSA, a set of some well-known state-of-the-art algorithms is considered for a significant comparison. Along with the basic GSA \cite{ras}, this set contains six algorithms namely particle swarm optimization (PSO)\cite{kennedy1995particle}, artificial bee colony (ABC) \cite{karaboga2007powerful}, differential evolution (DE) \cite{storn1997differential}, biogeography-based optimization (BBO) \cite{simon2008biogeography}, sine cosine algorithm (SCA) \cite{mirjalili2016sca}, and salp swarm algorithm (SSA) \cite{mirjalili2017salp}.
This comparison is conducted over Testbed $1$ with the following experimental setting:
\subsubsection{Experimental setting for Testbed 1}\label{exp_testbed1}
\begin{itemize}
\item To reduce the random discrepancy, $30$ independent runs have been conducted.
\item  population size $N=50$
\item Total number of iterations $T=1000$
\item Dimension $n=30$
\item The parameters of the considered algorithms are demonstrated in Table \ref{table:parameters}.
\end{itemize}

Since the proposed algorithm is a novel variant of GSA, therefore a fair comparison with some efficient and recent GSA variants is needed. For this, the proposed variant is compared with the basic GSA along with four robust GSA variants like PSOGSA \cite{mirjalili2010new}, GGSA \cite{mirjalili2014adaptive}, FVGGSA \cite{bansal2018fitness}, and PTGSA \cite{joshi2020parameter} over Testbed $2$ with the experimental setting recommended in CEC guidelines as follows: 

\subsubsection{Experimental setting for Testbed 2}\label{setting_CEC}
\begin{itemize}
\item Number of independent runs=$51$,
\item  population size $N=50$
\item Dimension $n=30$
\item Total number of iterations $T=6000$
\item The parameters of the considered GSA variants are demonstrated in Table \ref{table:parameters2}.  
\end{itemize}

\begin{center}
\begin{small}
\setlength{\LTleft}{-20cm plus -1fill}
\setlength{\LTright}{\LTleft}
\begin{longtable}{p{2cm}p{9cm}}
 \caption{Parameters of the considered algorithms over Testbed $1$}
 \label{table:parameters}
\endfirsthead
\caption* {\textbf{Table \ref{table:parameters} Continued:}}
\endhead
\toprule
 \textbf{Algorithm} &   \textbf{Parameters}  \\
 \midrule
ABC   & Food source=25, limit=25 $\times$ n\\
BBO  & Mutation probability = 0.01; number of elites = 2 \\
DE & $F_{min}$=0.2, $F_{max}$=0.8, CR=0.2 \\
PSO & $c_1$=$c_2$=2, inertia weight=1, inertia weight damping ratio=0.9 \\ 
SCA &  a = 2, $r_1$ deceases linearly from 2 to 0, $r_2$=$2 \times \pi \times rand()$, $r_3$=$2 \times rand()$, $r_4$ = rand()\\ 
SSA & $c_1$ is a exponentially decreasing function from 2 to 0, $c_2$ = rand() and $c_3$ = rand()
\\
GSA &  $\alpha$=20, $G_0$=100
\\
 COGSA &   $G_0$=100 and $\alpha$ is calculated from Algorithm \ref{algorithm:Chaos_G} \\

\bottomrule
\end{longtable}
 \end{small}
 \end{center}

\begin{center}
\begin{small}
\setlength{\LTleft}{-20cm plus -1fill}
\setlength{\LTright}{\LTleft}
\begin{longtable}{p{2cm}p{8cm}}
 \caption{Parameters of the considered algorithms over Testbed $2$}
 \label{table:parameters2}
\endfirsthead
\caption* {\textbf{Table \ref{table:parameters2} Continued:}}
\endhead
\toprule
 \textbf{Algorithm} &   \textbf{Parameters}  \\
 \midrule
 GSA    &  $\alpha$=20, $G_0$=100\\
 PSOGSA & $c_1$=0.5, $c_1$=1.5, $w$=rand(), $G_0$=1, $\alpha$=20\\ 
 GGSA & $G_0$=1, $\alpha$=20\\ 
 FVGGSA  & $\alpha$=10, variable $G_0$\\
 PTGSA & $G_0$=100, $\alpha \in [5, 70]$ \\
 COGSA & $G_0$=100, $\alpha$=1/3 $\times$ ($\alpha$ obtained by Algorithm \ref{algorithm:Chaos_G}) \\

\bottomrule
\end{longtable}
 \end{small}
 \end{center}

\subsection{Result and statistical analysis of experiments}
\subsubsection{Testbed 1}

\begin{figure}[tb!]
  \begin{subfigure}{0.5\textwidth}
    \includegraphics[height=4.5cm, width=8.3cm]{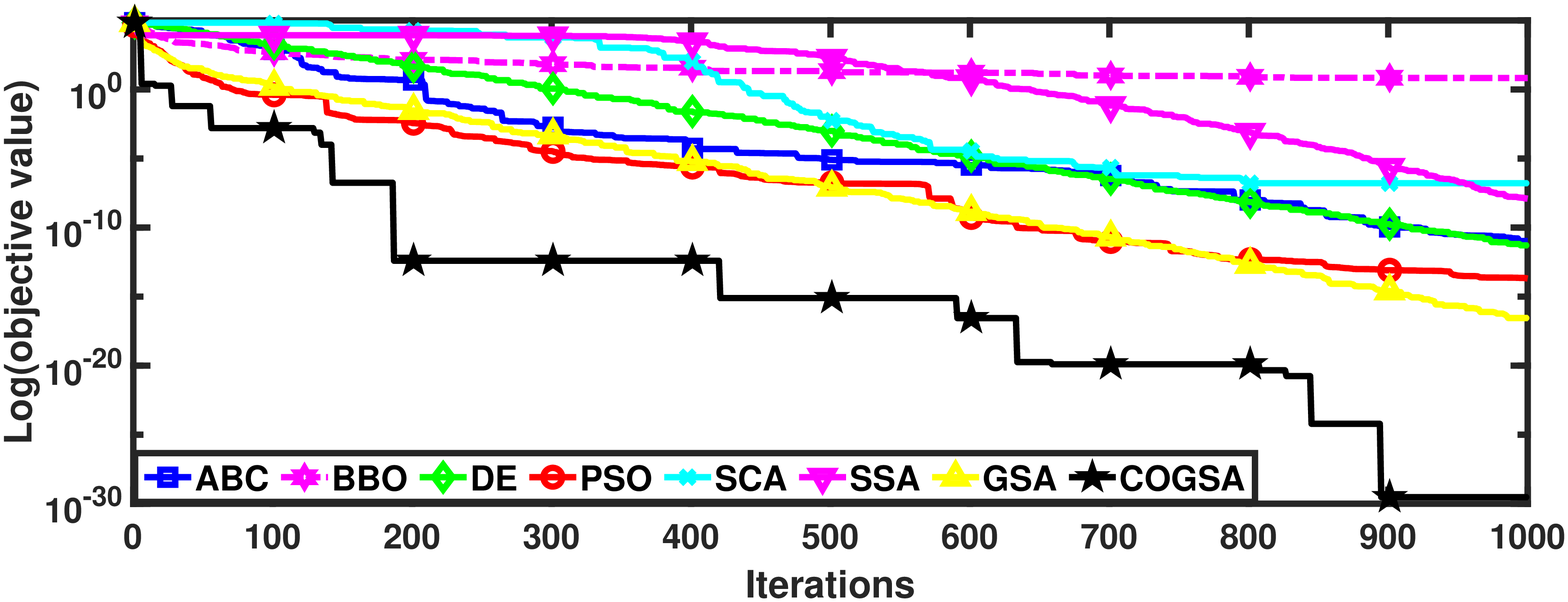}
    \caption{Convergence graphs for $f_1$}
  \end{subfigure}\hfill
   \begin{subfigure}{0.5\textwidth}
    \includegraphics[height=4.5cm, width=8.3cm]{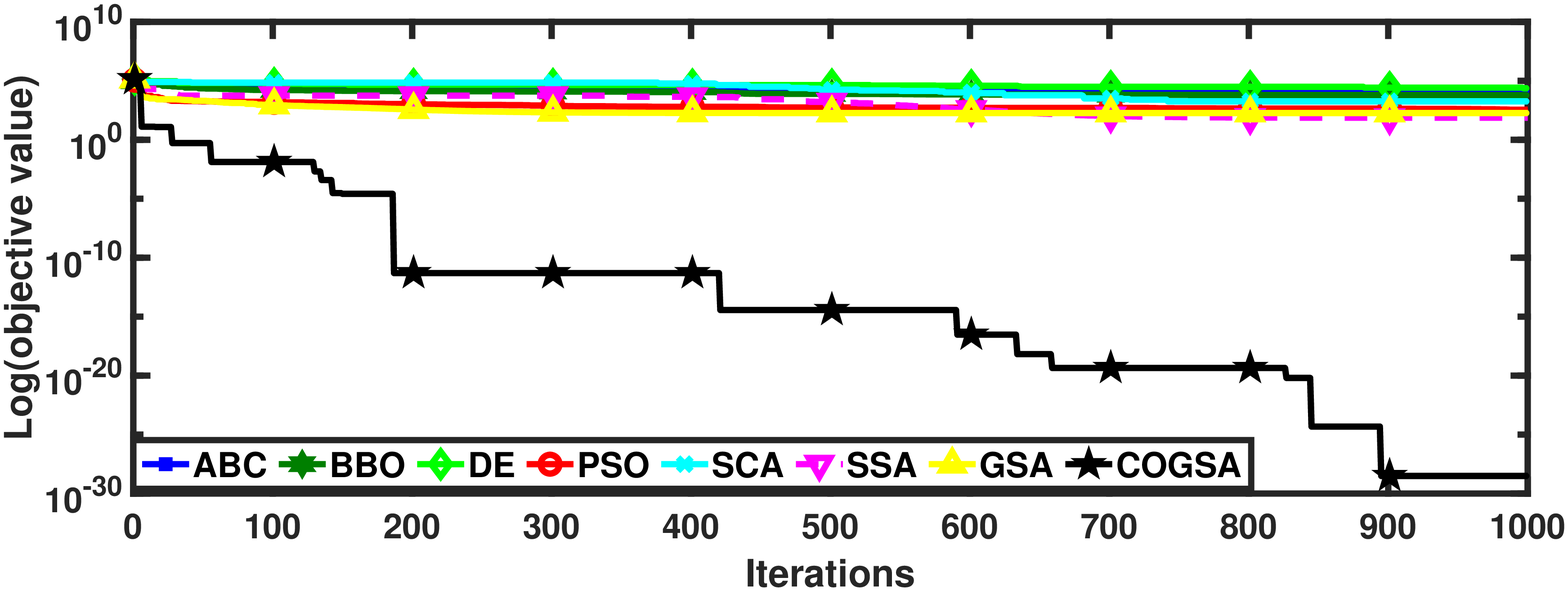}
    \caption{Convergence graphs for $f_3$}
  \end{subfigure}\hfill
    \begin{subfigure}{0.5\textwidth}
    \includegraphics[height=4.5cm, width=8.3cm]{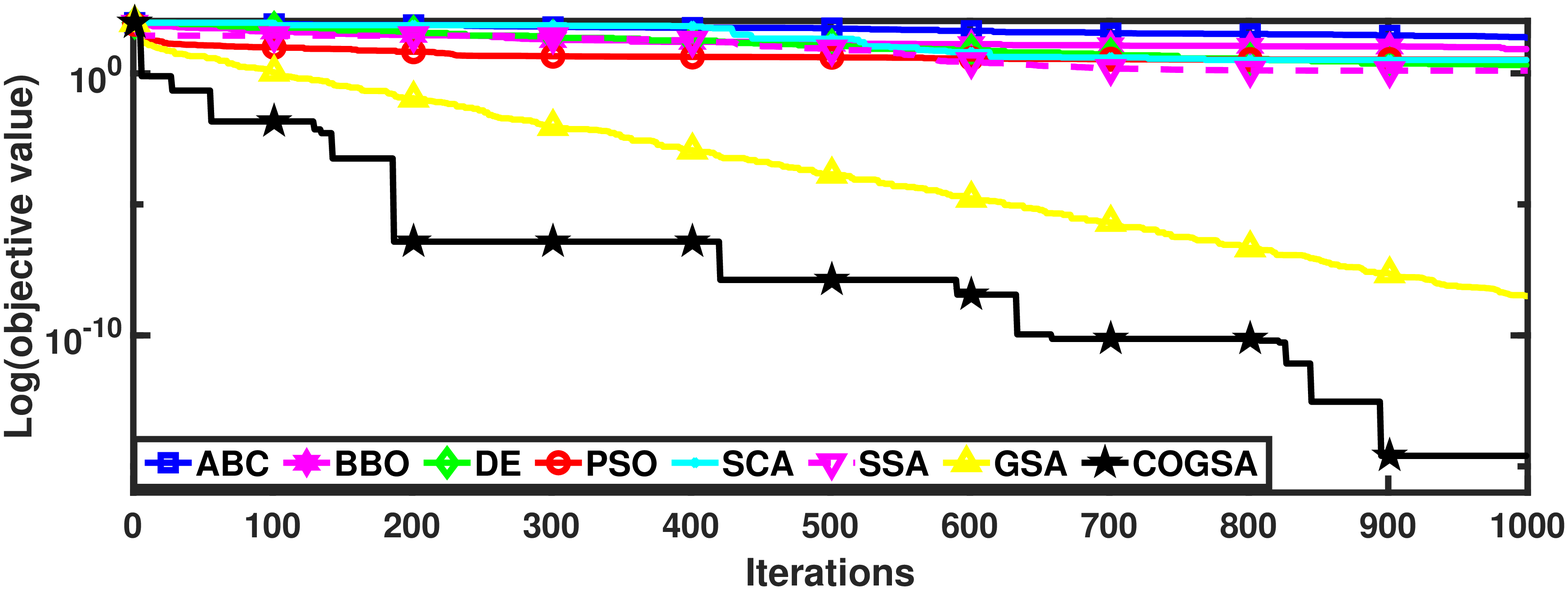}
    \caption{Convergence graphs for $f_4$}
  \end{subfigure}\hfill
  \begin{subfigure}{0.5\textwidth}
    \includegraphics[height=4.5cm, width=8.3cm]{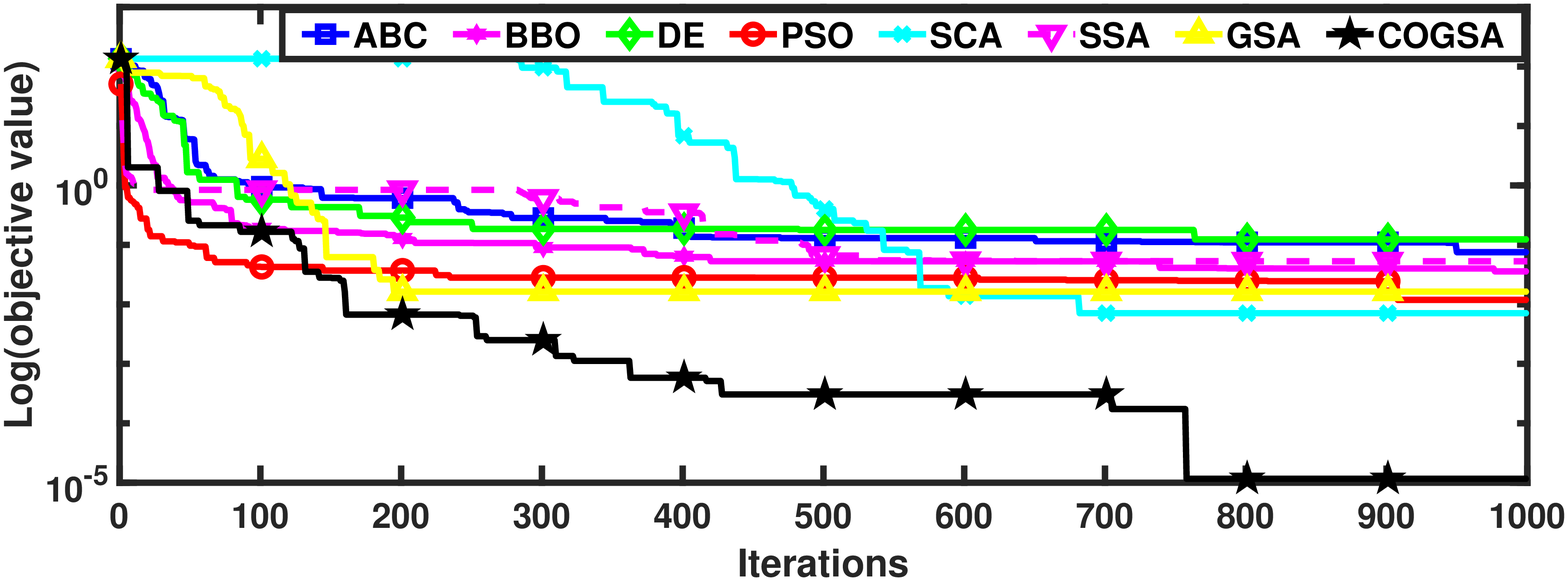}
    \caption{Convergence graphs for $f_7$}
  \end{subfigure}\hfill
  \begin{subfigure}{0.5\textwidth}
    \includegraphics[height=4.5cm, width=8.3cm]{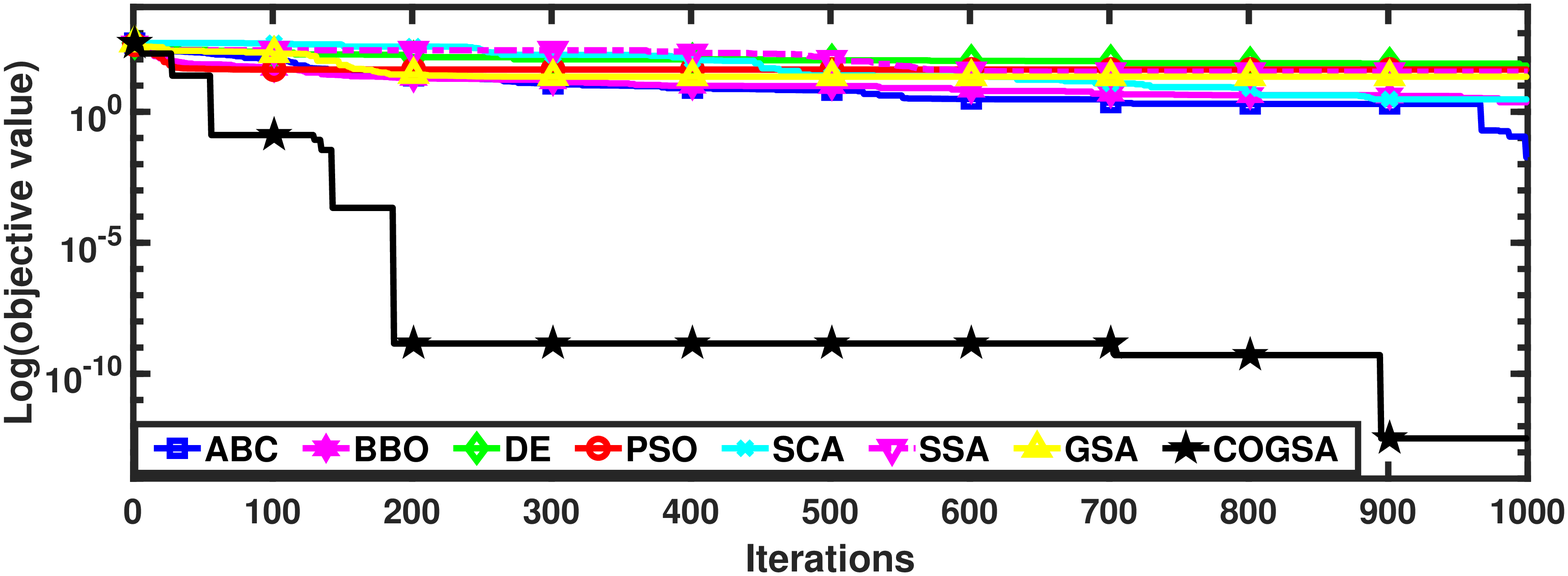}
    \caption{Convergence graphs for $f_{9}$}
    \end{subfigure}\hfill
   \begin{subfigure}{0.5\textwidth}
    \includegraphics[height=4.5cm, width=8.3cm]{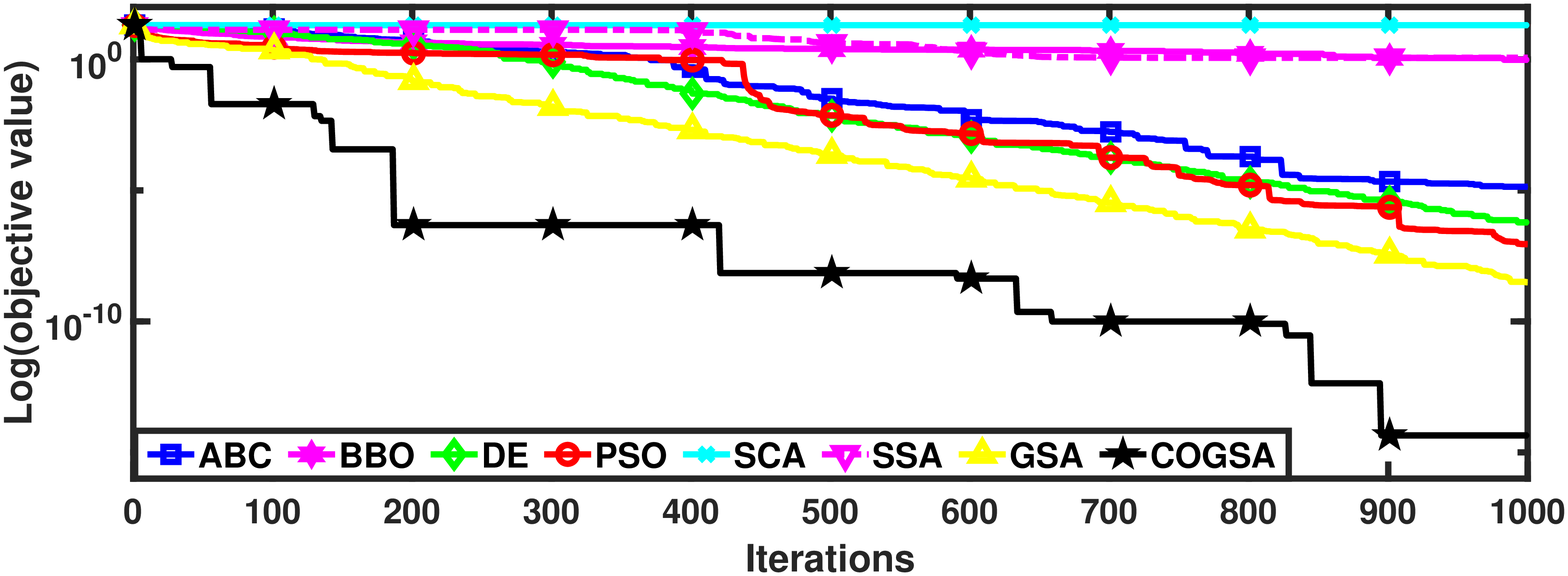}
    \caption{Convergence graphs for $f_{10}$}
    \end{subfigure}
  \caption{Convergence graphs of the considered algorithms} 
 \label{fig:con_graph_testbed1}
  \end{figure}

\begin{figure}[t]
  \begin{subfigure}{0.5\textwidth}
    \includegraphics[height=4.5cm, width=8.3cm]{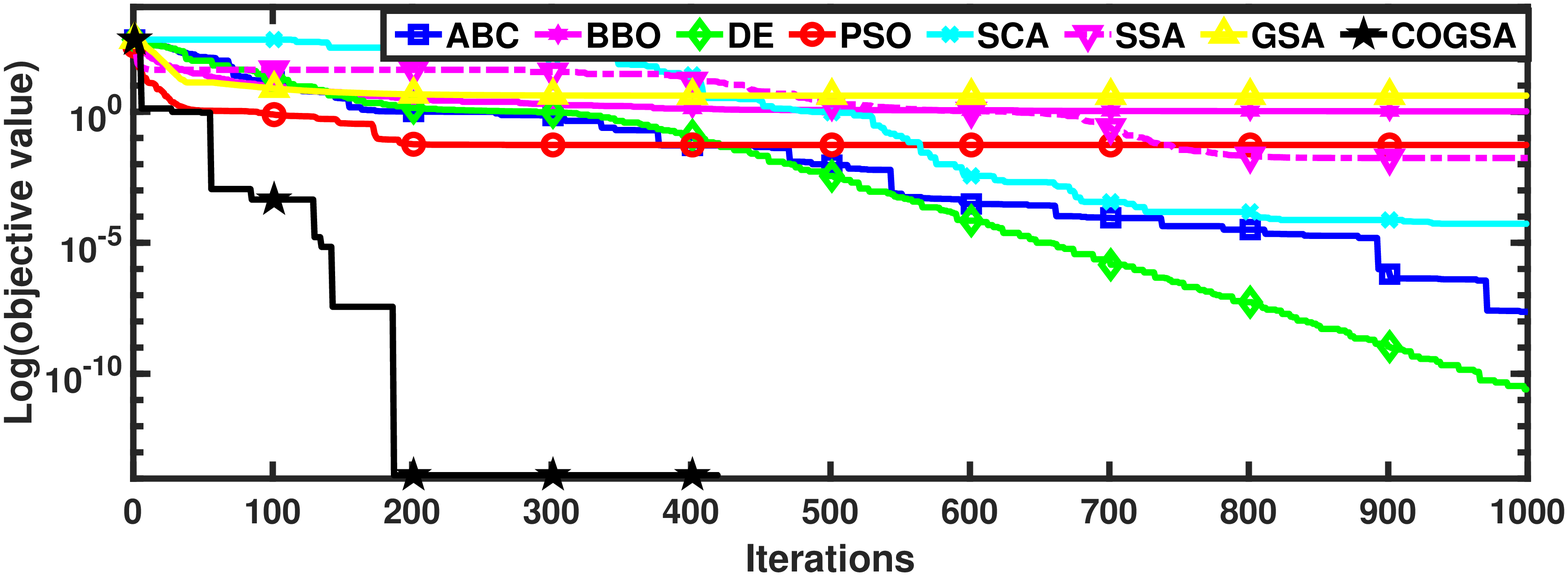}
    \caption{Convergence graphs for $f_{11}$}
  \end{subfigure}\hfill
   \begin{subfigure}{0.5\textwidth}
    \includegraphics[height=4.5cm, width=8.3cm]{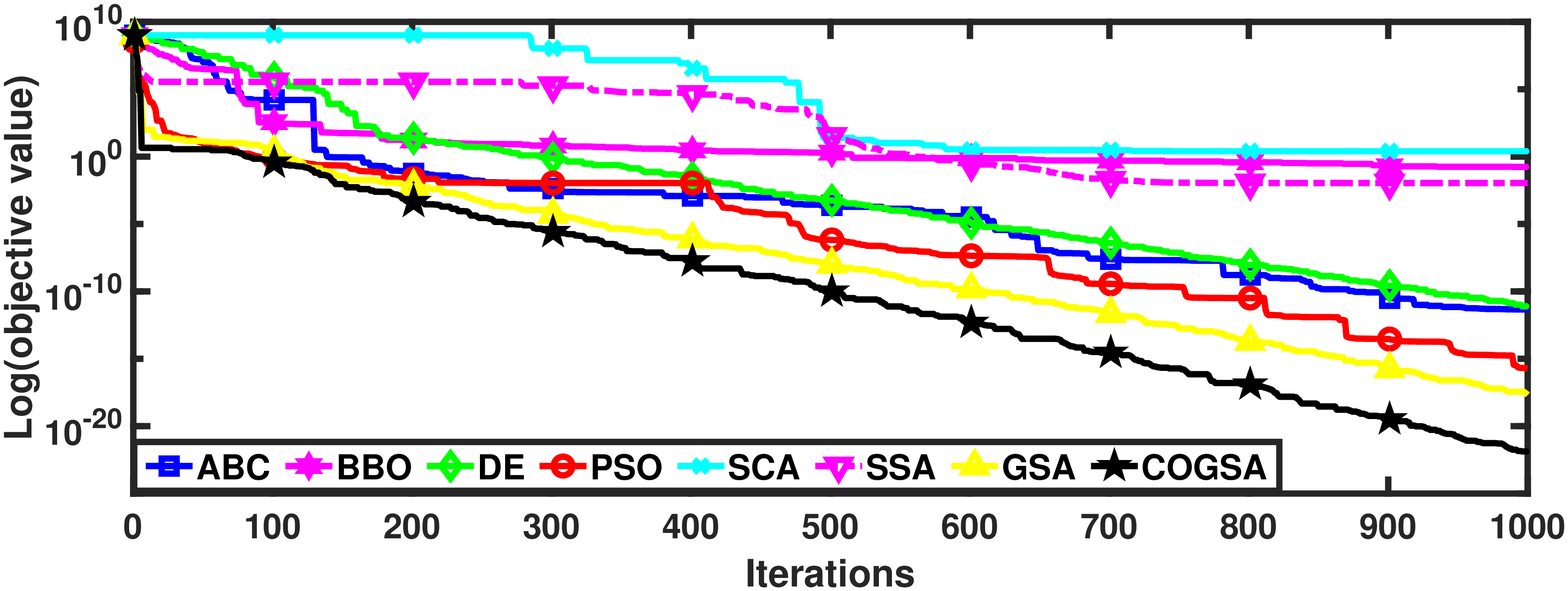}
    \caption{Convergence graphs for $f_{13}$}
  \end{subfigure}\hfill
    \begin{subfigure}{0.5\textwidth}
    \includegraphics[height=4.5cm, width=8.3cm]{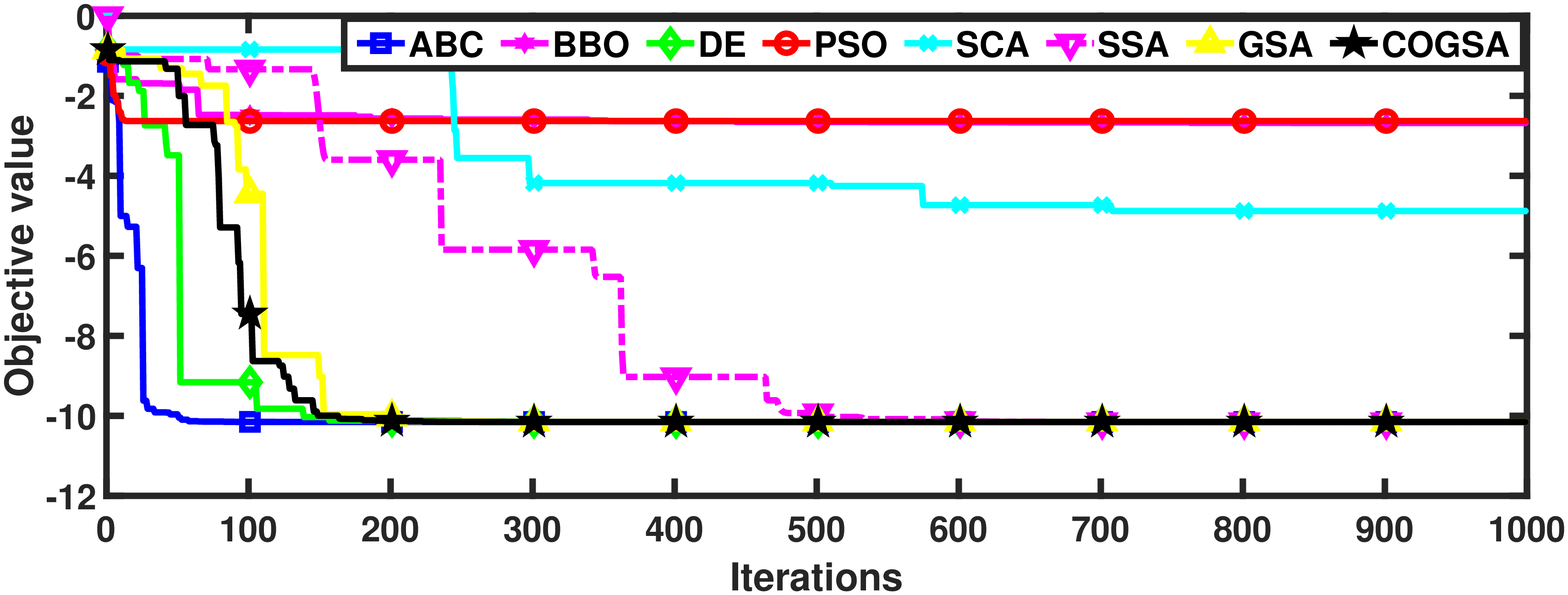}
    \caption{Convergence graphs for $f_{21}$}
  \end{subfigure}\hfill
  \begin{subfigure}{0.5\textwidth}
    \includegraphics[height=4.5cm, width=8.3cm]{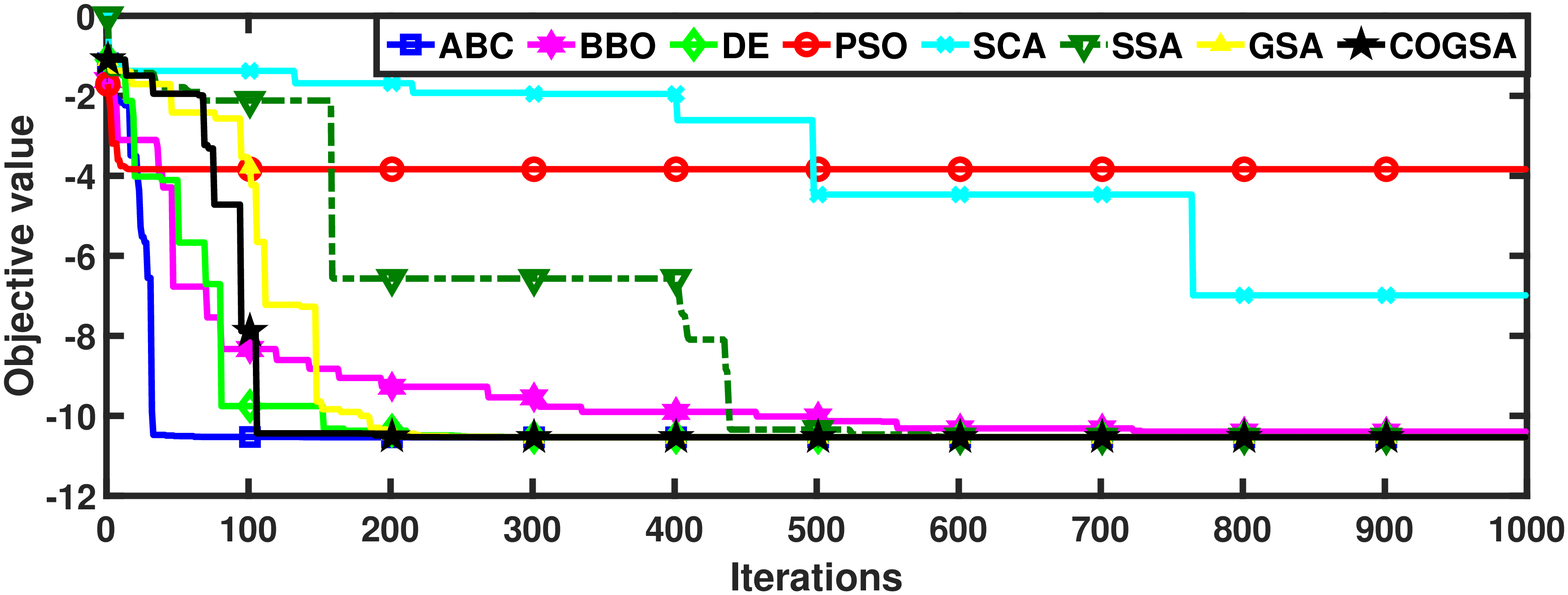}
    \caption{Convergence graphs for $f_{23}$}
        \end{subfigure}
  \caption{Convergence graphs of the considered algorithms } 
 \label{fig:con_graph_testbed11}
 \end{figure}

Table \ref{table:23testbed} demonstrates the experiment results of the proposed COGSA along with other considered algorithms over testbed $1$ followed by the experimental setting described in Section \ref{exp_testbed1}. These results are the mean value (Mean), best value (Best) and standard deviation (SD) of the optimal values obtained by the considered algorithms on $30$ runs. The best results are highlighted. As per the results in Table \ref{table:23testbed}, The proposed COGSA performs significantly well for $11$ test problems ($f_1$, $f_2$, $f_3$, $f_4$, $f_6$, $f_7$, $f_9$, $f_{10}$, $f_{11}$, $f_{13}$, and $f_{17}$) regarding all the three metric of comparison. In terms of mean value, COGSA achieves superiority for 17 test problems ($f_1$, $f_2$, $f_3$, $f_4$, $f_6$, $f_7$, $f_9$, $f_{10}$, $f_{11}$, $f_{13}$, $f_{16}$, $f_{17}$, $f_{18}$, $f_{19}$, $f_{20}$, $f_{22}$, and $f_{23}$). For $f_5$ and $f_{21}$, COGSA outperforms others except ABC. For $f_{12}$, COGSA performs superior than BBO, SCA, SSA and GSA. For $f_{15}$, COGSA is better than BBO and GSA. Under the best value consideration, COGSA dominates over $19$ test problems ($f_1$, $f_2$, $f_3$, $f_4$, $f_6$, $f_7$, $f_9$, $f_{10}$, $f_{11}$,  $f_{12}$, $f_{13}$, $f_{16}$, $f_{17}$, $f_{18}$, $f_{19}$, $f_{20}$, $f_{21}$, $f_{22}$, and $f_{23}$). As per the performance, the merits of the proposed COGSA can be pointed out as follows:

Out of $7$ unimodal problems ($f_1-f_7$), the supremacy over $6$ problems validates the remarkable exploitation ability of the proposed COGSA. While the supremacy over $4$ problems ($f_9$, $f_{10}$, $f_{11}$, and $f_{13}$) out of $6$ multimodal problems ($f_8-f_{13}$) confirms its efficient stagnation avoidance mechanism through which exploration gains more weight than exploitation to tackle the high number of local optima. Compared to multimodal problems, fixed dimensional multi modal problems ($f_{14}-f_{23}$) possess irregular fitness landscapes having less  local optima. A search that provides more weight to exploitation than exploration   
is required for these landscapes. The proposed COGSA has this specific search ability through which it finds the most promising regions for $7$ problems ($f_{16}$, $f_{17}$, $f_{18}$, $f_{19}$, $f_{20}$, $f_{22}$, and $f_{23}$) out of $10$ problems ($f_{14}-f_{23}$). Now, the performance differences of COGSA over other considered algorithms are examined statistically by the well-known Friedman test. It is a non-parametric test that calculates significant differences pairwise. In this study, this test is used at 1\% level of significance with the null hypothesis, `There is no significant difference between the results obtained by the considered pair'. Table \ref{table:Friedman} presents the p-values for the pairwise comparison between COGSA and other considered algorithms through the  post-hoc test procedure, namely `bonferroni'. As per the results shown in Table \ref{table:Friedman}, out of $161$ pairwise comparisons, $142$ p-values are less than $0.01$ which implies that there is a significant difference between those pairs. In order to investigate the performance of the proposed COGSA over high dimensional search space, a scalability test is conducted and compared with basic GSA. For this, $13$ scalable problems ($f_1-f_{13}$) of Testbed $1$ are reconsidered with higher dimensions as $n=50$ and $n=100$. Both the algorithms follow the same experimental setting which is described in Section \ref{exp_testbed1}. It is evident in the results shown in Table \ref{table:scalability} that the proposed COGSA significantly outperforms in the higher dimensions as well. On the other hand, basic GSA performs drastically poorly as dimension increases. To further verify the local search abilities of the considered algorithms, convergence graphs are plotted over four unimodal problems ($f_1$, $f_3$, $f_4$, and $f_7$), four multi-modal problems ($f_9$, $f_{10}$, $f_{11}$, and $f_{13}$) and two fixed dimension multimodal problems ($f_{21}$ and $f_{23}$). These graphs are shown in Figures \ref{fig:con_graph_testbed1}-\ref{fig:con_graph_testbed11}. These figures assure the fast converge rate of the proposed COGSA. The Chaos embedded OBL strategy provides an efficient stagnation avoidance mechanism through which the algorithm effectively supervises the local optima regions and leads the search towards global optima regions. The above comparison and analyses conclude that the proposed COGSA has a robust, fast, and diversified search mechanism that significantly deals with unimodal, multimodal, fixed multimodal, and high dimensional problems.

\begin{center}
\begin{footnotesize}
\setlength{\LTleft}{-20cm plus -1fill}
\setlength{\LTright}{\LTleft}

\setlength\tabcolsep{2pt}
\begin{longtable}{cccccccccc}

 \caption{Experimental results over Testbed 1}
 \label{table:23testbed}
\endfirsthead
\caption* {\textbf{Table \ref{table:23testbed} Continued:}}
\endhead
\toprule[0.9pt]
\textbf{TP } & \textbf{metrics} &  \textbf{ABC}&  \textbf{BBO}  &	\textbf{DE} &	\textbf{PSO} & \textbf{SCA} &	\textbf{SSA} &	\textbf{GSA}&	\textbf{COGSA} \\
 \midrule[0.9pt]

	\multirow{3}{*} {$f_{1}$}			&	Mean	&	1.5861E-11	&	5.5720E+00	&	6.2884E-12	&	2.1138E-14	&	2.1693E-03	&	9.2502E-09	&	2.3404E-17	&	\textbf{5.1898E-30}	\\	&	Best	&	1.6075E-12	&	2.8370E+00	&	2.3946E-12	&	3.5184E-22	&	5.0209E-07	&	4.6753E-09	&	1.1451E-17	&	\textbf{1.5724E-30}	\\	&	SD	&	1.8604E-11	&	1.3678E+00	&	2.6402E-12	&	7.2874E-14	&	9.8414E-03	&	1.7716E-09	&	7.0499E-18	&	\textbf{2.4398E-30}	\\ [1ex]
		
	\multirow{3}{*} {$f_{2}$}			&	Mean	&	8.5811E-07	&	8.4164E-01	&	5.2882E-08	&	9.7108E-03	&	4.3358E-06	&	4.8992E-01	&	2.2989E-08	&	\textbf{1.0638E-14}	\\	&	Best	&	4.3889E-07	&	5.2546E-01	&	2.9388E-08	&	3.9178E-10	&	1.0463E-08	&	8.5002E-05	&	1.5741E-08	&	\textbf{6.5627E-15}	\\	&	SD	&	3.4071E-07	&	1.4096E-01	&	1.4327E-08	&	2.8694E-02	&	7.1872E-06	&	7.3100E-01	&	3.6023E-09	&	\textbf{2.2802E-15}	\\ [1ex]
		
	\multirow{3}{*} {$f_{3}$}			&	Mean	&	1.1105E+04	&	9.0942E+03	&	2.6935E+04	&	3.8575E+02	&	2.6583E+03	&	4.2282E+01	&	2.4909E+02	&	\textbf{6.5905E-29}	\\	&	Best	&	5.8713E+03	&	4.0696E+03	&	1.6536E+04	&	1.7006E+02	&	1.1389E+02	&	8.4786E+00	&	1.2213E+02	&	\textbf{6.9411E-30}	\\	&	SD	&	2.2742E+03	&	2.3636E+03	&	4.2234E+03	&	1.4672E+02	&	3.2358E+03	&	3.1769E+01	&	1.0007E+02	&	\textbf{7.2348E-29}	\\ [1ex]	
	
	\multirow{3}{*} {$f_{4}$}			&	Mean	&	1.9808E+01	&	6.2053E+00	&	2.1407E+00	&	3.7537E+00	&	1.4604E+01	&	4.6223E+00	&	3.5668E-09	&	\textbf{1.1556E-15}	\\	&	Best	&	8.9166E+00	&	3.8890E+00	&	1.3318E+00	&	2.3650E+00	&	2.6184E+00	&	7.5282E-01	&	2.3469E-09	&	\textbf{6.3685E-16}	\\	&	SD	&	4.1055E+00	&	1.0318E+00	&	3.5456E-01	&	6.0507E-01	&	9.4135E+00	&	3.2005E+00	&	5.6638E-10	&	\textbf{5.1681E-16}	\\ [1ex]
		
	\multirow{3}{*} {$f_{5}$}			&	Mean	&	\textbf{1.8860E+00}	&	3.6996E+02	&	3.6236E+01	&	7.4323E+01	&	5.8041E+01	&	6.4859E+01	&	4.6122E+01	&	2.7075E+01	\\	&	Best	&	9.4606E-02	&	1.9151E+02	&	2.5087E+01	&	\textbf{9.2027E+00}	&	2.8104E+01	&	2.2873E+01	&	2.5752E+01	&	2.6654E+01	\\	&	SD	&	1.6604E+00	&	3.0148E+02	&	1.2252E+01	&	4.7644E+01	&	7.9371E+01	&	7.1081E+01	&	4.5338E+01	&	\textbf{1.8334E-01}	\\ [1ex]
		
	\multirow{3}{*} {$f_{6}$}			&	Mean	&	0.0000E+00	&	7.4333E+00	&	5.3333E-01	&	4.0000E-01	&	0.0000E+00	&	8.5333E+00	&	0.0000E+00	&	\textbf{0.0000E+00}	\\	&	Best	&	0.0000E+00	&	1.0000E+00	&	0.0000E+00	&	0.0000E+00	&	0.0000E+00	&	2.0000E+00	&	0.0000E+00	&	\textbf{0.0000E+00}	\\	&	SD	&	0.0000E+00	&	3.1271E+00	&	4.9889E-01	&	4.8990E-01	&	0.0000E+00	&	3.5752E+00	&	0.0000E+00	&	\textbf{0.0000E+00}	\\ [1ex]

	\multirow{3}{*} {$f_{7}$}			&	Mean	&	1.1192E-01	&	2.0845E-02	&	1.6091E-01	&	1.3640E-02	&	2.3734E-02	&	4.9972E-02	&	1.8481E-02	&	\textbf{4.9403E-05}	\\	&	Best	&	7.3880E-02	&	1.1169E-02	&	9.9618E-02	&	8.6290E-03	&	2.8997E-03	&	2.6445E-02	&	8.2748E-03	&	\textbf{3.0793E-06}	\\	&	SD	&	2.3650E-02	&	8.0522E-03	&	2.8687E-02	&	3.9494E-03	&	1.3951E-02	&	1.2486E-02	&	6.0221E-03	&	\textbf{4.2668E-05}	\\ [1ex]	
	
\bottomrule[0.9pt]	
\pagebreak	
	
\toprule[0.9pt]
\textbf{TP } & \textbf{metrics} &  \textbf{ABC}&  \textbf{BBO}  &	\textbf{DE} &	\textbf{PSO} & \textbf{SCA} &	\textbf{SSA} &	\textbf{GSA}&	\textbf{COGSA} \\
 \midrule[0.9pt]

	\multirow{3}{*} {$f_{8}$}			&	Mean	&	-1.2213E+04	&	\textbf{-1.2554E+04}	&	-1.2499E+04	&	-6.2444E+03	&	-4.0603E+03	&	-7.6023E+03	&	-2.8782E+03	&	-2.7145E+03	\\	&	Best	&	-1.2451E+04	&	-1.2562E+04	&	\textbf{-1.2569E+04}	&	-8.9767E+03	&	-4.7418E+03	&	-8.8567E+03	&	-3.6279E+03	&	-3.8395E+03	\\	&	SD	&	1.3452E+02	&	\textbf{4.4783E+00}	&	2.8483E+02	&	8.5778E+02	&	2.4803E+02	&	7.5013E+02	&	3.3885E+02	&	4.5213E+02	\\ [1ex]

	\multirow{3}{*} {$f_{9}$}			&	Mean	&	4.5050E-01	&	2.3131E+00	&	6.1090E+01	&	4.5304E+01	&	1.5272E+01	&	4.7260E+01	&	1.5356E+01	&	\textbf{7.5791E-15}	\\	&	Best	&	3.7089E-09	&	1.1807E+00	&	4.8425E+01	&	1.9899E+01	&	3.6771E-07	&	2.8854E+01	&	7.9597E+00	&	\textbf{0.0000E+00}	\\	&	SD	&	6.2849E-01	&	8.1027E-01	&	6.3362E+00	&	1.8714E+01	&	2.1537E+01	&	1.1422E+01	&	4.5062E+00	&	\textbf{4.0815E-14}	\\ [1ex]

	\multirow{3}{*} {$f_{10}$}			&	Mean	&	1.3151E-05	&	1.2243E+00	&	6.2074E-07	&	4.2009E-02	&	1.1359E+01	&	1.4252E+00	&	3.5742E-09	&	\textbf{4.4409E-15}	\\	&	Best	&	3.0962E-06	&	6.6914E-01	&	3.4539E-07	&	9.2755E-10	&	6.7980E-05	&	1.9342E-05	&	2.7045E-09	&	\textbf{4.4409E-15}	\\	&	SD	&	8.0820E-06	&	2.2240E-01	&	1.3835E-07	&	2.2542E-01	&	9.5701E+00	&	8.0505E-01	&	5.0068E-10	&	\textbf{0.0000E+00}	\\ [1ex]

	\multirow{3}{*} {$f_{11}$}			&	Mean	&	1.8231E-03	&	1.0518E+00	&	1.4227E-10	&	3.5583E-02	&	1.7645E-01	&	1.0171E-02	&	3.9241E+00	&	\textbf{7.4015E-18}	\\	&	Best	&	4.3396E-12	&	1.0166E+00	&	1.0138E-11	&	6.6613E-16	&	1.8607E-05	&	2.5940E-08	&	1.6952E+00	&	\textbf{0.0000E+00}	\\	&	SD	&	4.9324E-03	&	1.9953E-02	&	3.0001E-10	&	3.8695E-02	&	2.3967E-01	&	1.0910E-02	&	1.8692E+00	&	\textbf{3.9858E-17}	\\ [1ex]

	\multirow{3}{*} {$f_{12}$}			&	Mean	&	1.2390E-12	&	3.9285E-02	&	\textbf{6.6757E-13}	&	6.9113E-03	&	1.6173E+00	&	3.0668E+00	&	2.6389E-02	&	1.3823E-02	\\	&	Best	&	9.6924E-14	&	6.6375E-03	&	1.7022E-13	&	1.4222E-21	&	3.2539E-01	&	1.9660E-01	&	7.5197E-20	&	\textbf{3.8117E-24}	\\	&	SD	&	1.8654E-12	&	4.2005E-02	&	\textbf{3.0871E-13}	&	2.5860E-02	&	2.5570E+00	&	1.6404E+00	&	4.4232E-02	&	3.5241E-02	\\ [1ex]

	\multirow{3}{*} {$f_{13}$}			&	Mean	&	2.2286E-11	&	2.7652E-01	&	3.8520E-12	&	2.5637E-03	&	2.9034E+00	&	1.9479E+00	&	2.2344E-18	&	\textbf{1.0618E-22}	\\	&	Best	&	6.8182E-13	&	1.4850E-01	&	9.5099E-13	&	2.7479E-19	&	1.9747E+00	&	4.9698E-10	&	9.5452E-19	&	\textbf{4.8191E-23}	\\	&	SD	&	5.9273E-11	&	8.5370E-02	&	2.4630E-12	&	8.3596E-03	&	1.0326E+00	&	1.0449E+01	&	6.9537E-19	&	\textbf{3.3334E-23}	\\ [1ex]

	\multirow{3}{*} {$f_{14}$}			&	Mean	&	9.9800E-01	&	\textbf{9.9931E-01}	&	1.0311E+00	&	2.4758E+00	&	1.2626E+00	&	9.9800E-01	&	3.7632E+00	&	4.7747E+00	\\	&	Best	&	\textbf{9.9800E-01}	&	\textbf{9.9800E-01}	&	\textbf{9.9800E-01}	&	\textbf{9.9800E-01}	&	\textbf{9.9800E-01}	&	\textbf{9.9800E-01}	&	9.9819E-01	&	1.0354E+00	\\	&	SD	&	\textbf{1.4895E-16}	&	4.5358E-03	&	1.7843E-01	&	2.4071E+00	&	6.7444E-01	&	2.3464E-16	&	2.2081E+00	&	3.0952E+00	\\ [1ex]

	\multirow{3}{*} {$f_{15}$}			&	Mean	&	\textbf{6.4824E-04}	&	1.1544E-02	&	6.9497E-04	&	1.0273E-03	&	8.0532E-04	&	7.8285E-04	&	2.2138E-03	&	1.9290E-03	\\	&	Best	&	3.8946E-04	&	7.5607E-04	&	4.9483E-04	&	\textbf{3.0749E-04}	&	3.5114E-04	&	3.4503E-04	&	7.9921E-04	&	1.0882E-03	\\	&	SD	&	1.6800E-04	&	9.9749E-03	&	\textbf{1.1644E-04}	&	3.5979E-03	&	3.4205E-04	&	2.2785E-04	&	1.2376E-03	&	5.5262E-04	\\ [1ex]

	\multirow{3}{*} {$f_{16}$}			&	Mean	&	\textbf{-1.0316E+00}	&	-1.0299E+00	&	\textbf{-1.0316E+00}	&	\textbf{-1.0316E+00}	&	\textbf{-1.0316E+00}	&	\textbf{-1.0316E+00}	&	\textbf{-1.0316E+00}	&	\textbf{-1.0316E+00}	\\	&	Best	&	\textbf{-1.0316E+00}	&	\textbf{-1.0316E+00}	&	\textbf{-1.0316E+00}	&	\textbf{-1.0316E+00}	&	\textbf{-1.0316E+00}	&	\textbf{-1.0316E+00}	&	\textbf{-1.0316E+00}	&	\textbf{-1.0316E+00}	\\	&	SD	&	\textbf{4.9651E-16}	&	1.4162E-03	&	5.5880E-16	&	6.5994E-16	&	1.1207E-05	&	3.3361E-15	&	5.7332E-16	&	6.6613E-16	\\ [1ex]

	\multirow{3}{*} {$f_{17}$}			&	Mean	&\textbf{3.9789E-01}	&	4.0260E-01	&	\textbf{3.9789E-01}	&	\textbf{3.9789E-01}	&	3.9853E-01	&	\textbf{3.9789E-01}	&	\textbf{3.9789E-01}	&	\textbf{3.9789E-01}	\\	&	Best	&	\textbf{3.9789E-01}	&	3.9798E-01	&	\textbf{3.9789E-01}	&	\textbf{3.9789E-01}	&	3.9796E-01	&	\textbf{3.9789E-01}	&	\textbf{3.9789E-01}	&	\textbf{3.9789E-01}	\\	&	SD	&	\textbf{0.0000E+00}	&	7.3010E-03	&	8.7030E-16	&	\textbf{0.0000E+00}	&	5.6084E-04	&	2.0693E-14	&	\textbf{0.0000E+00}	&	\textbf{0.0000E+00}	\\ [1ex]

\bottomrule[0.9pt]	
\pagebreak	
	
\toprule[0.9pt]
\textbf{TP } & \textbf{metrics} &  \textbf{ABC}&  \textbf{BBO}  &	\textbf{DE} &	\textbf{PSO} & \textbf{SCA} &	\textbf{SSA} &	\textbf{GSA}&	\textbf{COGSA} \\
 \midrule[0.9pt]

	\multirow{3}{*} {$f_{18}$}			&	Mean	&	\textbf{3.0000E+00}	&	4.8423E+00	&	\textbf{3.0000E+00}	&	\textbf{3.0000E+00}	&	\textbf{3.0000E+00}	&	\textbf{3.0000E+00}	&	\textbf{3.0000E+00}	&	\textbf{3.0000E+00}	\\	&	Best	&	\textbf{3.0000E+00}	&	3.0004E+00	&	\textbf{3.0000E+00}	&	\textbf{3.0000E+00}	&	\textbf{3.0000E+00}	&	\textbf{3.0000E+00}	&	\textbf{3.0000E+00}	&	\textbf{3.0000E+00}	\\	&	SD	&	9.9825E-05	&	6.7885E+00	&	\textbf{5.3167E-16}	&	7.5626E-16	&	8.8463E-06	&	5.5752E-14	&	1.9860E-15	&	1.0127E-15	\\ [1ex]

	\multirow{3}{*} {$f_{19}$}			&	Mean	&	\textbf{-3.8628E+00}	&	-3.8626E+00	&	\textbf{-3.8628E+00}	&	\textbf{-3.8628E+00}	&	-3.8555E+00	&	\textbf{-3.8628E+00}	&	\textbf{-3.8628E+00}	&	\textbf{-3.8628E+00}	\\	&	Best	&	\textbf{-3.8628E+00}	&	\textbf{-3.8628E+00}	&	\textbf{-3.8628E+00}	&	\textbf{-3.8628E+00}	&	-3.8612E+00	&	\textbf{-3.8628E+00}	&	\textbf{-3.8628E+00}	&	\textbf{-3.8628E+00}	\\	&	SD	&	\textbf{2.2367E-15}	&	1.5376E-04	&	2.4964E-15	&	2.6509E-15	&	2.2056E-03	&	3.2059E-14	&	2.3929E-15	&	2.6645E-15	\\ [1ex]

	\multirow{3}{*} {$f_{20}$}			&	Mean	&	\textbf{-3.3220E+00	}&	-3.2623E+00	&	\textbf{-3.3202E+00}	&	-3.2744E+00	&	-2.9829E+00	&	-3.2300E+00	&	\textbf{-3.3220E+00}	&	\textbf{-3.3220E+00}	\\	&	Best	&	\textbf{-3.3220E+00}	&	\textbf{-3.3220E+00}	&	\textbf{-3.3220E+00}	&	\textbf{-3.3220E+00}	&	-3.1878E+00	&	\textbf{-3.3220E+00}	&	\textbf{-3.3220E+00}	&	\textbf{-3.3220E+00}	\\	&	SD	&	\textbf{1.2482E-15}	&	5.9485E-02	&	9.8001E-03	&	5.8245E-02	&	2.8307E-01	&	5.0778E-02	&	1.3323E-15	&	1.3323E-15	\\ [1ex]

	\multirow{3}{*} {$f_{21}$}			&	Mean	&	\textbf{-1.0153E+01}	&	-5.0284E+00	&	-1.0152E+01	&	-6.3099E+00	&	-3.6862E+00	&	-8.8168E+00	&	-7.1126E+00	&	-1.0093E+01	\\	&	Best	&	\textbf{-1.0153E+01}	&	-1.0133E+01	&	-1.0153E+01	&	-1.0153E+01	&	-6.1709E+00	&	-1.0153E+01	&	-1.0153E+01	&	\textbf{-1.0153E+01}	\\	&	SD	&	\textbf{1.1003E-14}	&	3.1554E+00	&	6.5080E-03	&	3.4601E+00	&	1.8280E+00	&	2.7220E+00	&	3.3157E+00	&	1.1066E-01	\\ [1ex]

	\multirow{3}{*} {$f_{22}$}			&	Mean	&	\textbf{-1.0403E+01}	&	-6.2628E+00	&	-1.0227E+01	&	-6.1468E+00	&	-4.8450E+00	&	-9.6196E+00	&	\textbf{-1.0403E+01}	&	\textbf{-1.0403E+01}	\\	&	Best	&	\textbf{-1.0403E+01}	&	-1.0388E+01	&	\textbf{-1.0403E+01}	&	\textbf{-1.0403E+01}	&	-8.0366E+00	&	\textbf{-1.0403E+01}	&	\textbf{-1.0403E+01}	&	\textbf{-1.0403E+01}	\\	&	SD	&	3.5706E-05	&	3.5717E+00	&	9.4689E-01	&	3.5416E+00	&	1.5829E+00	&	2.0314E+00	&	\textbf{5.6173E-16}	&	9.7295E-16	\\ [1ex]

	\multirow{3}{*} {$f_{23}$}			&	Mean	&	\textbf{-1.0536E+01}	&	-6.4988E+00	&	\textbf{-1.0536E+01}	&	-6.6427E+00	&	-4.2651E+00	&	-9.3277E+00	&	\textbf{-1.0536E+01}	&	\textbf{-1.0536E+01}	\\	&	Best	&	\textbf{-1.0536E+01}	&	-1.0525E+01	&	\textbf{-1.0536E+01}	&	\textbf{-1.0536E+01}	&	-8.1818E+00	&	\textbf{-1.0536E+01}	&	\textbf{-1.0536E+01}	&	\textbf{-1.0536E+01}	\\	&	SD	&	1.8257E-03	&	3.7379E+00	&	\textbf{1.6852E-15}	&	3.7292E+00	&	1.7010E+00	&	2.7290E+00	&	1.8346E-15	&	2.0254E-15	\\ [1ex]

\bottomrule[0.9pt]
\end{longtable}

\end{footnotesize}
 \end{center} 

\begin{center}
\begin{small}
\setlength{\LTleft}{-20cm plus -1fill}
\setlength{\LTright}{\LTleft}
\begin{longtable}{p{1.2cm}p{1.6cm}p{1.6cm}p{1.6cm}p{1.6cm}p{1.6cm}p{1.6cm}p{1.6cm}}
 \caption{p-Values for comparison of COGSA with the considered algorithms}
 \label{table:Friedman}
\endfirsthead
\caption* {\textbf{Table \ref{table:Friedman} Continued:}}
\endhead
\toprule[0.9pt]
 \textbf{Problem} &  \textbf{ABC}&  \textbf{BBO}  &	\textbf{DE} &	\textbf{PSO} & \textbf{SCA}  & \textbf{SSA} & \textbf{GSA}	\\
 \midrule[0.9pt]
	
$f_1$	& $<2E-16$ & $<2E-16$ & $<2E-16$ & $<2E-16$ & $<2E-16$ & $<2E-16$ & $<2E-16$ \\[1ex]
$f_2$	& $<2E-16$ & $<2E-16$ & $<2E-16$ & $<2E-16$ & $<2E-16$ & $<2E-16$ & $<2.2E-11$ \\[1ex]	
 $f_3$	& $<2E-16$ & $<2E-16$ & $<2E-16$ & $<2E-16$ & $<2E-16$ & $5.9E-13$ & $<2E-16$ \\[1ex]	
  $f_4$	& $<2E-16$ & $<2E-16$ & $<2E-16$ & $<2E-16$ & $<2E-16$ & $5.9E-13$ & $6.0E-07$ \\[1ex]
  $f_5$	& $<2E-16$ & $<2E-16$ & $1.0E-08$ & $4.3E-11$ & $3.5E-13$ & $0.00331$ & $1$ \\[1ex]
 		 			
 $f_6$	& $1$ & $<2E-16$ & $<2E-16$ & $4.3E-13$ & $1$ & $<2E-16$ & $1$ \\[1ex]

\bottomrule[0.9pt]	
\pagebreak	
	
\toprule[0.9pt]
 \textbf{Problem} &  \textbf{ABC}&  \textbf{BBO}  &	\textbf{DE} &	\textbf{PSO} & \textbf{SCA}  & \textbf{SSA} & \textbf{GSA}	\\
 \midrule[0.9pt]

$f_7$	& $<2E-16$ & $<2E-16$ & $<2E-16$ & $5.0E-14$ & $<2E-16$ & $<2E-16$ & $<2E-16$ \\[1ex]

 $f_8$	& $<2E-16$ & $<2E-16$ & $<2E-16$ & $<2E-16$ & $<2E-16$ & $<2E-16$ & $1$ \\[1ex]			
 	$f_9$	& $4.6E-13$ & $<2E-16$ & $<2E-16$ & $<2E-16$ & $<2E-16$ & $<2E-16$ & $<2E-16$ \\[1ex]	
 	
 	$f_{10}$	& $<2E-16$ & $<2E-16$ & $<2E-16$ & $<2E-16$ & $<2E-16$ & $<2E-16$ & $ 4.9E-13$ \\[1ex]
 	
 	$f_{11}$	& $<2E-16$ & $<2E-16$ & $8.2E-14$ & $<2E-16$ & $<2E-16$ & $<2E-16$ & $<2E-16$ \\[1ex]	
 	$f_{12}$	& $<2E-16$ & $<2E-16$ & $<2E-16$ & $8.6E-07$ & $<2E-16$ & $<2E-16$ & $1.1E-13$ \\[1ex]
 	
$f_{13}$	& $<2E-16$ & $<2E-16$ & $<2E-16$ & $<2E-16$ & $<2E-16$ & $<2E-16$ & $1.1E-09$ \\[1ex] 
$f_{14}$	& $<2E-16$ & $<2E-16$ & $<2E-16$ & $<2E-16$ & $<2E-16$ & $<2E-16$ & $1$ \\[1ex]	
$f_{15}$	& $<2E-16$ & $4.7E-08 $ & $<2E-16$ & $<2E-16$ & $<2E-16$ & $<2E-16$ & $1$ \\[1ex]

$f_{16}$	& $<2E-16$ & $<2E-16$ & $<2E-16$ & $1$ & $<2E-16$ & $<2E-16$ & $6.8E-13$ \\[1ex]
 $f_{17}$	& $1$ & $<2E-16$ & $1.3E-13$ & $1$ & $<2E-16$ & $<2E-16$ & $1$ \\[1ex]		
 		
$f_{18}$	& $<2E-16$ & $<2E-16$ & $1$ & $0.022$ & $<2E-16$ & $<2E-16$ & $<2E-16$ \\[1ex] 	$f_{19}$	& $<2E-16$ & $<2E-16$ & $ 2.7E-11$ & $1$ & $<2E-16$ & $<2E-16$ & $<2E-16$ \\[1ex]	

$f_{20}$	& $1.9E-11$ & $<2E-16$ & $6.6E-10$ & $7.8E-16$ & $<2E-16$ & $<2E-16$ & $1$ \\[1ex]
 	
$f_{21}$	& $1$ & $<2E-16$ & $1$ & $ 4.5E-10$ & $<2E-16$ & $1.4E-15$ & $1.4E-08$ \\[1ex] 	
$f_{22}$	& $1.7E-11$ & $<2E-16$ & $ 1.7E-11$ & $<2E-16$ & $<2E-16$ & $<2E-16$ & $0.38$ \\[1ex]	
 	
$f_{23}$	& $< 2e-16$ & $<2E-16$ & $ 0.02972$ & $<2E-16$ & $<2E-16$ & $<2E-16$ & $ 7.7e-09 $ \\[1ex]

\bottomrule[0.9pt]
\end{longtable}
 \end{small}
 \end{center}

\begin{center}
\begin{table}[t]

 \caption{Scalability test} 
\label{table:scalability}
\scalebox{0.72}{
\begin{tabular}
{cccccccccccc} \toprule[0.9pt] \multirow{2}{*} {\textbf{Dim}} & \multirow{2}{*}{\textbf{Algorithm}} & \multicolumn{2}{c}{$f_1$}&  \multicolumn{2}{c}{$f_2$} & \multicolumn{2}{c}{$f_3$} & \multicolumn{2}{c}{$f_4$} & \multicolumn{2}{c}{$f_5$}\\ \cmidrule[0.9pt]{3-4}\cmidrule[0.9pt]{5-6}\cmidrule[0.9pt]{7-8} \cmidrule[0.9pt]{9-10} \cmidrule[0.9pt]{11-12}

& & \textbf{Mean} & \textbf{SD}  & \textbf{Mean} & \textbf{SD}  & \textbf{Mean} & \textbf{SD} & \textbf{Mean} & \textbf{SD} & \textbf{Mean} & \textbf{SD} \\ \midrule[0.9pt] 

50	&	GSA	&	6.82E-17	&	2.20E-17	&	5.59E-08	&	1.21E-08	&	9.90E+02	&	3.42E+02	&	3.98E+00	&	1.39E+00	&	5.27E+01	&	1.97E+01	\\
	&	COGSA	&	\textbf{5.40E-30}	&	\textbf{2.08E-30}	&	\textbf{1.40E-14}	&	\textbf{2.76E-15}	&	\textbf{7.67E-29}	&	\textbf{8.98E-29}	&	\textbf{1.03E-15}	&	\textbf{2.95E-16}	&	\textbf{4.71E+01}	&	\textbf{1.91E-01}	\\
100	&	GSA	&	6.85E+01	&	6.54E+01	&	1.06E+00	&	5.55E-01	&	4.93E+03	&	1.09E+03	&	1.04E+01	&	1.17E+00	&	1.91E+03	&	1.28E+03	\\
	&	COGSA	&	\textbf{7.25E-30}	&	\textbf{3.52E-30}	&	\textbf{2.01E-14}	&	\textbf{5.24E-15}	&	\textbf{1.36E-28}	&	\textbf{1.11E-28}	&	\textbf{8.81E-16}	&	\textbf{3.54E-16}	&	\textbf{9.71E+01}	&	\textbf{2.49E-01}	\\[1ex]
	
\toprule[0.9pt] \multirow{2}{*} {\textbf{Dim}} & \multirow{2}{*}{\textbf{Algorithm}} & \multicolumn{2}{c}{$f_6$}&  \multicolumn{2}{c}{$f_7$} & \multicolumn{2}{c}{$f_8$} & \multicolumn{2}{c}{$f_9$} & \multicolumn{2}{c}{$f_{10}$}\\ \cmidrule[0.9pt]{3-4}\cmidrule[0.9pt]{5-6}\cmidrule[0.9pt]{7-8} \cmidrule[0.9pt]{9-10} \cmidrule[0.9pt]{11-12}  

& & \textbf{Mean} & \textbf{SD}  & \textbf{Mean} & \textbf{SD}  & \textbf{Mean} & \textbf{SD} & \textbf{Mean} & \textbf{SD} & \textbf{Mean} & \textbf{SD} \\ \midrule[0.9pt] 

50	&	GSA	&	8.33E-01	&	1.21E+00	&	5.77E-02	&	1.55E-02	&	-3.65E+03	&	6.65E+02	&	3.04E+01	&	4.65E+00	&	4.79E-09	&	5.85E-10	\\
	&	COGSA	&	\textbf{0.00E+00}	&	\textbf{0.00E+00}	&	\textbf{3.28E-05}	&	\textbf{2.76E-05}	&	\textbf{-3.67E+03}	&	\textbf{4.05E+02}	&	\textbf{3.60E-14}	&	\textbf{1.84E-13}	&	\textbf{4.09E-15}	&	\textbf{1.07E-15}	\\
100	&	GSA	&	3.52E+02	&	1.74E+02	&	5.61E-01	&	1.91E-01	&	-4.68E+03	&	\textbf{6.05E+02}	&	7.86E+01	&	9.84E+00	&	1.08E+00	&	4.09E-01	\\
	&	COGSA	&	\textbf{0.00E+00}	&	\textbf{0.00E+00}	&	\textbf{3.91E-05}	&	\textbf{4.02E-05}	&	\textbf{-5.06E+03}	&	7.42E+02	&	\textbf{3.79E-15}	&	\textbf{2.04E-14}	&	\textbf{1.84E-15}	&	\textbf{1.57E-15}	\\

\toprule[0.9pt] \multirow{2}{*} {\textbf{Dim}} & \multirow{2}{*}{\textbf{Algorithms}} & \multicolumn{2}{c}{$f_{11}$}&  \multicolumn{2}{c}{$f_{12}$} & \multicolumn{2}{c}{$f_{13}$} & \multicolumn{2}{c}{} & \multicolumn{2}{c}{}\\ \cmidrule[0.9pt]{3-4}\cmidrule[0.9pt]{5-6}\cmidrule[0.9pt]{7-8} 

& & \textbf{Mean} & \textbf{SD}  & \textbf{Mean} & \textbf{SD}  & \textbf{Mean} & \textbf{SD} & \textbf{} & \textbf{} & \textbf{} & \textbf{} \\ \cmidrule[0.9pt]{1-8}
50	&	GSA	&	1.66E+01	&	4.70E+00	&	5.06E-01	&	3.49E-01	&	2.48E+00	&	3.14E+00	&		&		&		&		\\
	&	COGSA	&	\textbf{0.00E+00}	&	\textbf{0.00E+00}	&	\textbf{1.04E-02}	&	\textbf{2.32E-02}	&	\textbf{5.87E-03}	&	\textbf{1.79E-02}	&		&		&		&		\\
100	&	GSA	&	5.43E+01	&	7.32E+00	&	1.92E+00	&	3.58E-01	&	7.61E+01	&	1.58E+01	&		&		&		&		\\
	&	COGSA	&	\textbf{2.48E-13}	&	\textbf{4.31E-13}	&	\textbf{1.35E-02}	&	\textbf{9.81E-03}	&	\textbf{9.68E-01}	&	\textbf{4.20E-01}	&		&		&		&		\\
\cmidrule[0.9pt]{1-8}
\end{tabular}
}
\end{table}
\end{center}

\subsubsection{Testbed 2}
\begin{figure}[tb!]
  \begin{subfigure}{0.5\textwidth}
    \includegraphics[height=4.5cm, width=8.3cm]{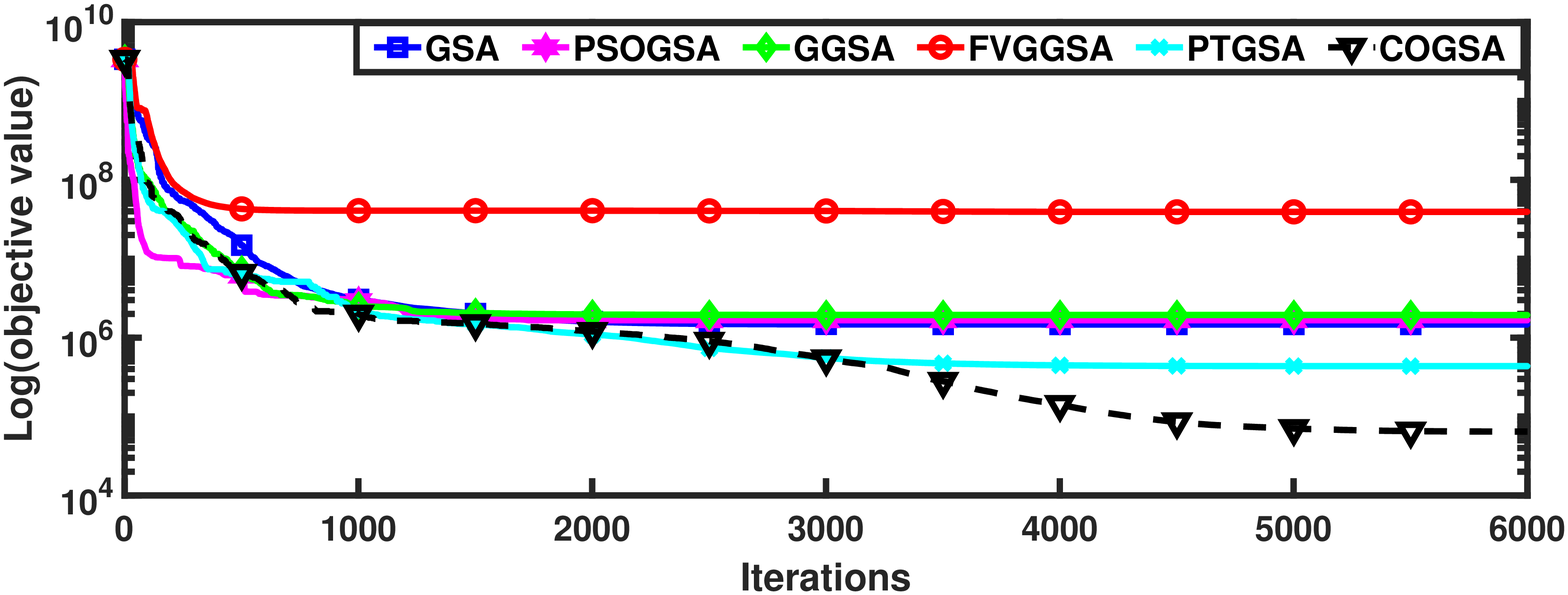}
    \caption{Convergence graphs for $g_1$}
  \end{subfigure}\hfill
    \begin{subfigure}{0.5\textwidth}
    \includegraphics[height=4.5cm, width=8.3cm]{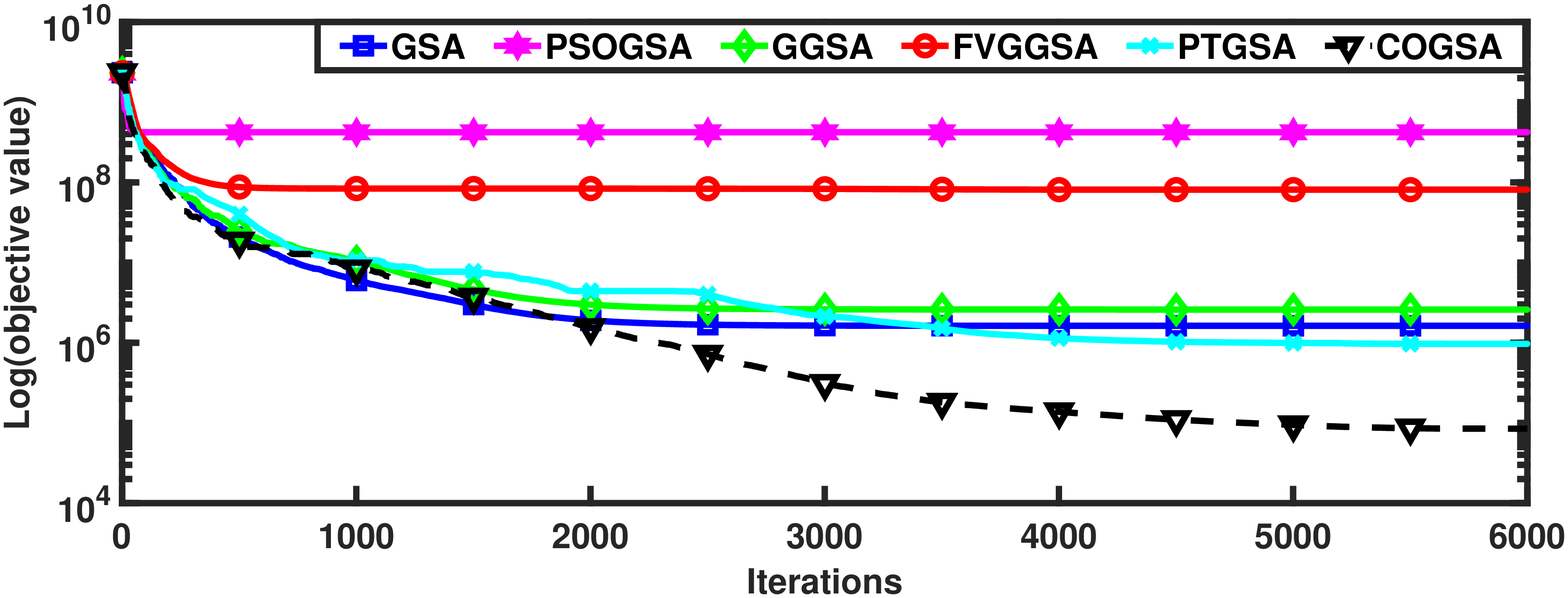}
    \caption{Convergence graphs for $g_3$}
  \end{subfigure}\hfill
   \begin{subfigure}{0.5\textwidth}
    \includegraphics[height=4.5cm, width=8.3cm]{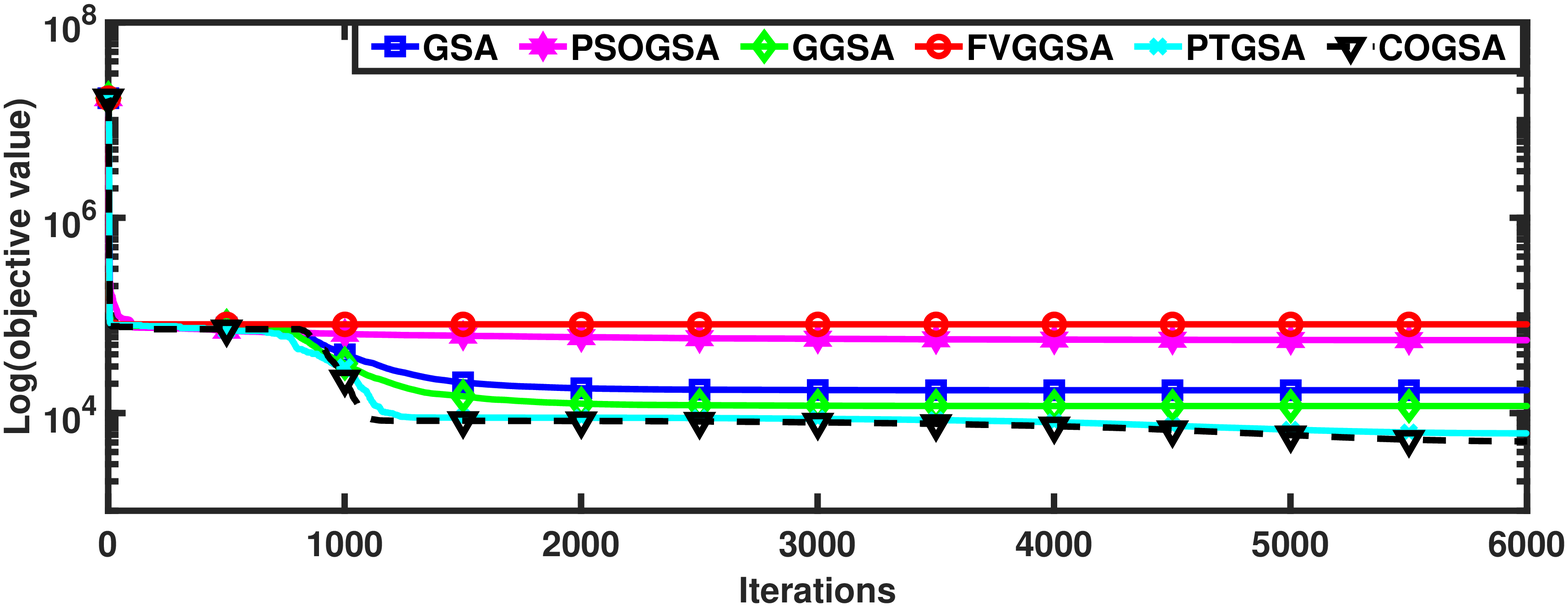}
    \caption{Convergence graphs for $g_{4}$}
    \end{subfigure}\hfill   
   \begin{subfigure}{0.5\textwidth}
    \includegraphics[height=4.5cm, width=8.3cm]{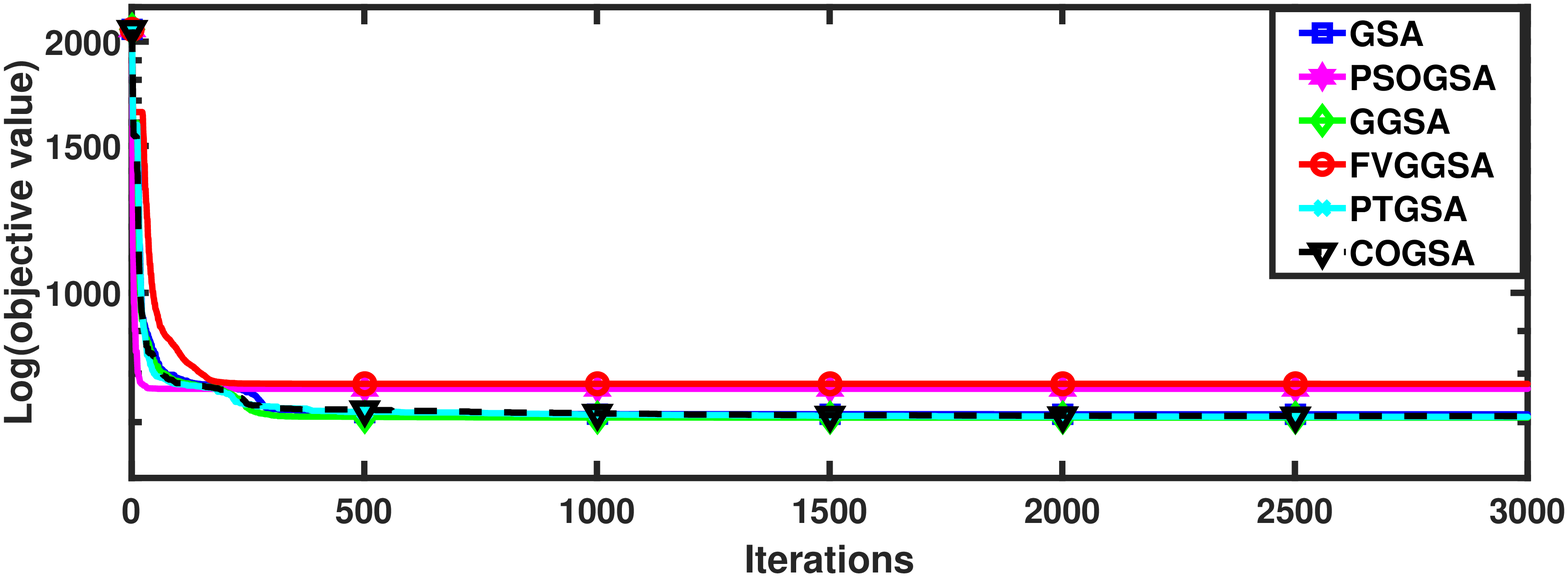}
    \caption{Convergence graphs for $g_{14}$}
  \end{subfigure}\hfill
   \begin{subfigure}{0.5\textwidth}
    \includegraphics[height=4.5cm, width=8.3cm]{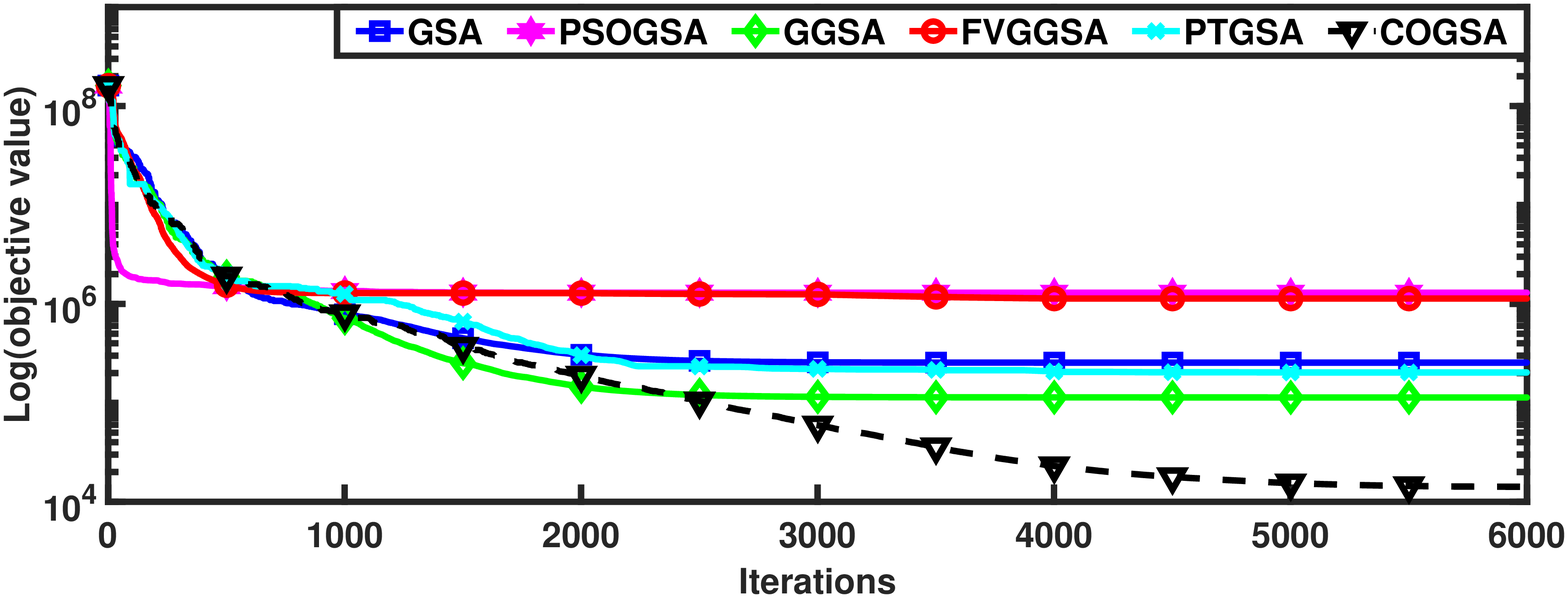}
    \caption{Convergence graphs for $g_{16}$}
    \end{subfigure}     
    \begin{subfigure}{0.5\textwidth}
    \includegraphics[height=4.5cm, width=8.3cm]{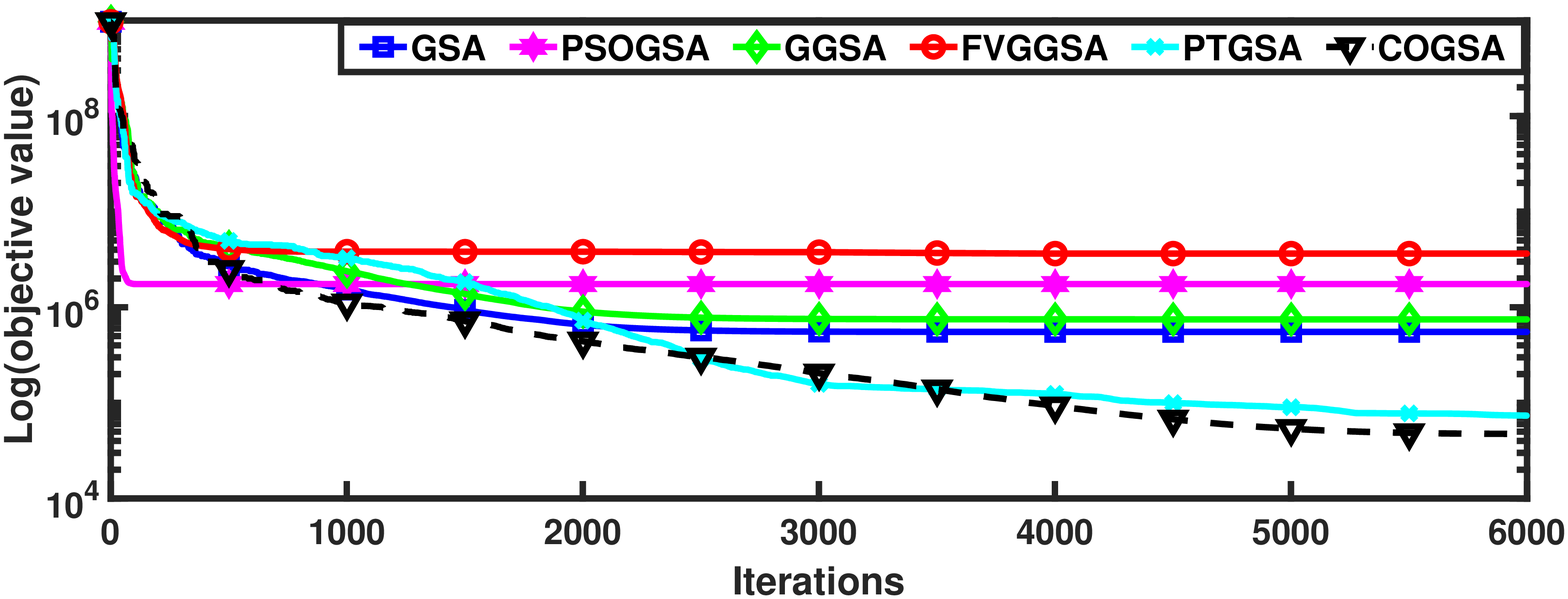}
    \caption{Convergence graphs for $g_{19}$}
  \end{subfigure}\hfill
  \caption{Convergence graphs of the considered algorithms} 
 \label{fig:converge_cec}
  \end{figure}

\begin{figure}[t]
  \begin{subfigure}{0.5\textwidth}
    \includegraphics[height=4.5cm, width=8.3cm]{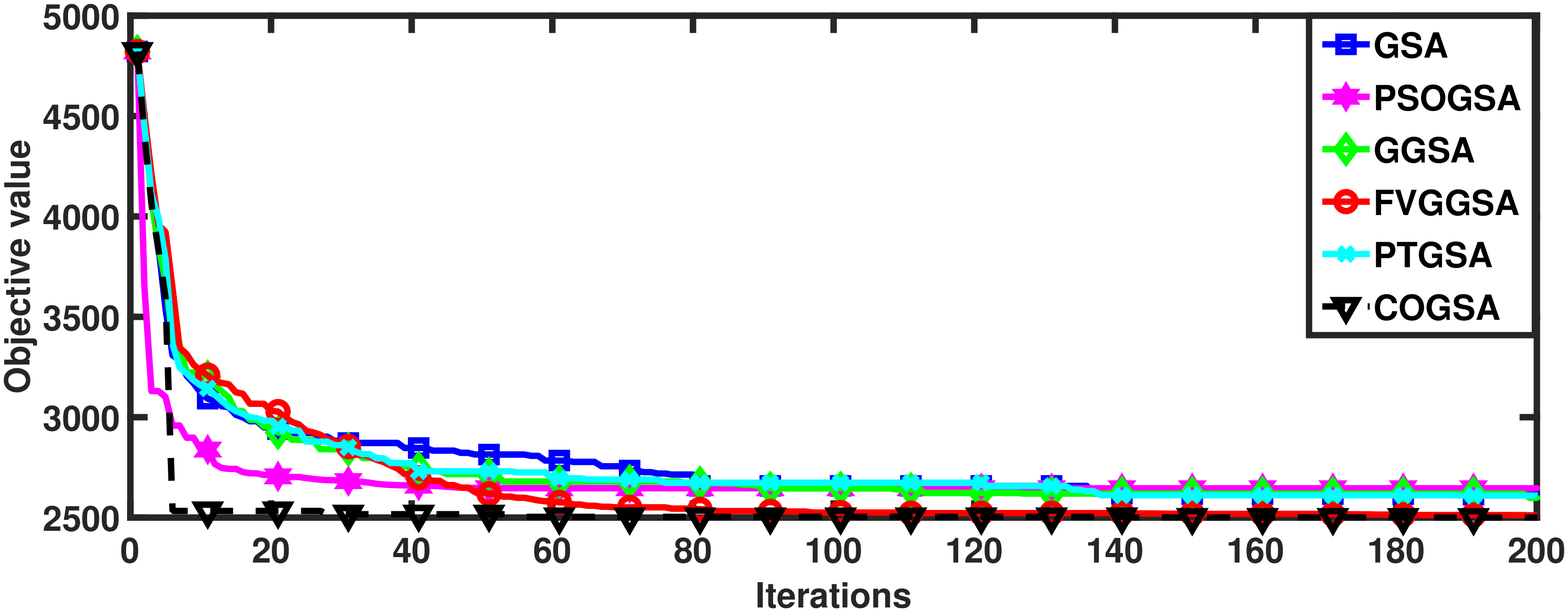}
    \caption{Convergence graphs for $g_{25}$}
  \end{subfigure}\hfill
  \begin{subfigure}{0.5\textwidth}
    \includegraphics[height=4.5cm, width=8.3cm]{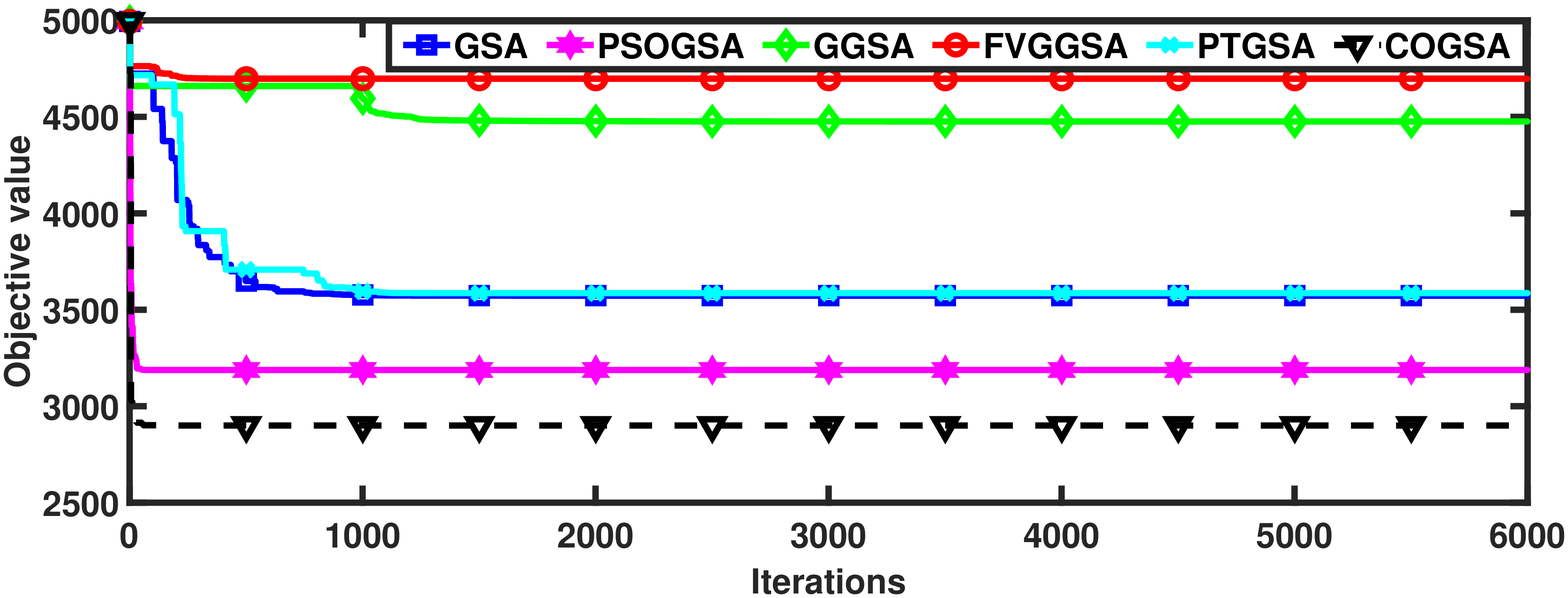}
    \caption{Convergence graphs for $g_{28}$}
    \end{subfigure}\hfill
   \begin{subfigure}{0.5\textwidth}
    \includegraphics[height=4.5cm, width=8.3cm]{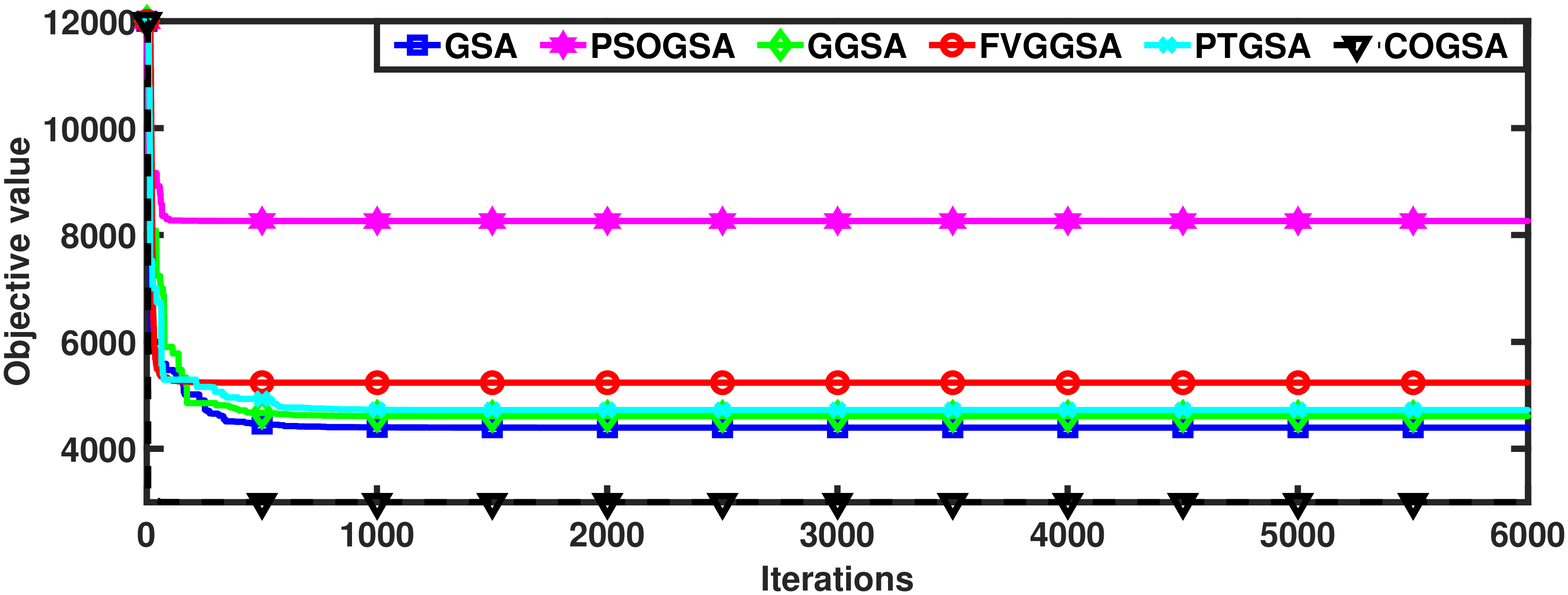}
    \caption{Convergence graphs for $g_{29}$}
    \end{subfigure}\hfill
  \begin{subfigure}{0.5\textwidth}
    \includegraphics[height=4.5cm, width=8.3cm]{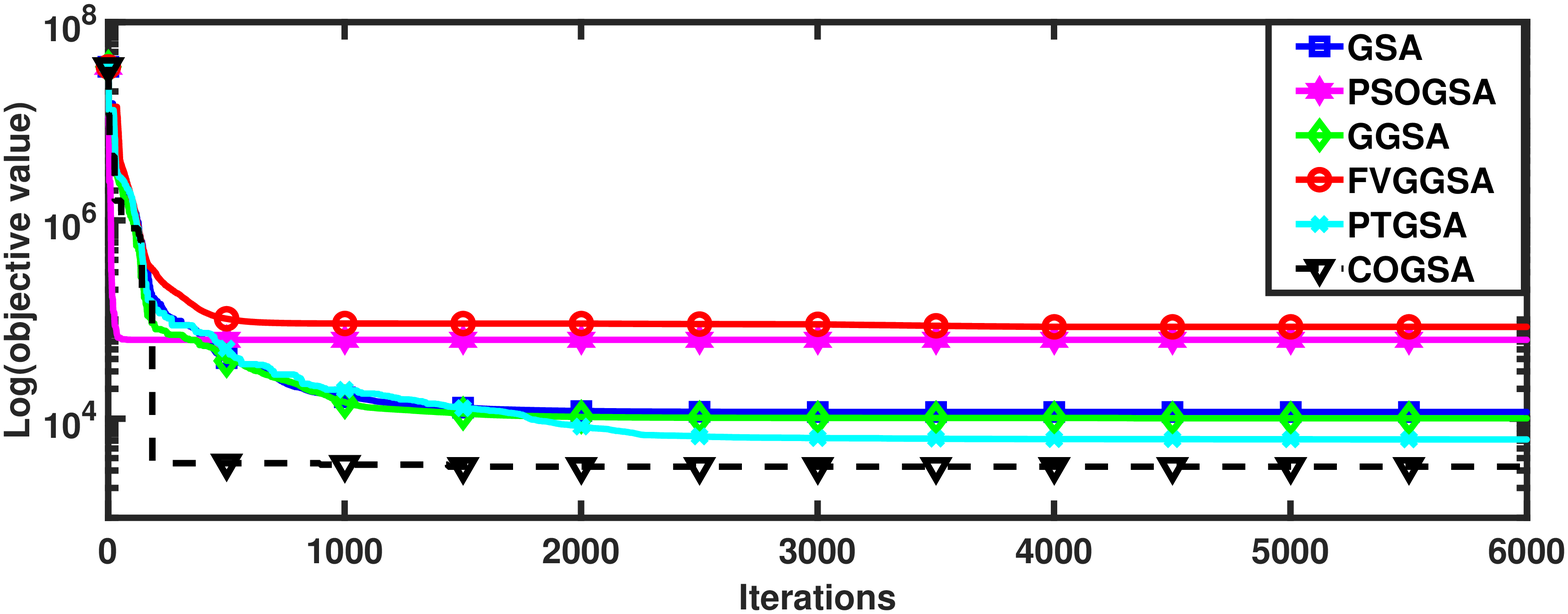}
    \caption{Convergence graphs for $g_{30}$}
    \end{subfigure}
  \caption{Convergence graphs of the considered algorithms} 
 \label{fig:converge_cec1}
  \end{figure}

To further evaluate the search abilities of the proposed COGSA, more rigid and challenging problems of Testbed $2$ are taken under consideration. Section \ref{setting_CEC} describes the parameter setting along with the common experimental setting of the proposed COGSA and all the other considered GSA variants. The performances of all the algorithms are based on fitness error and summarized in Table \ref{table:results_testbed2}. Based on fitness error, three metrics of comparison namely mean error (Mean), standard deviation of error (SD) and Friedman test (p-value) are listed in Table \ref{table:results_testbed2}. The level of significance and the null hypothesis for conducting the Friedman test over Testbed $2$ are the same as Testbed $1$. It is clear from Table \ref{table:results_testbed2} that COGSA achieves the optimal mean value for four unimodal ($g_1$, $g_2$, $g_3$, and $g_4$), six multimodal ($g_5$, $g_7$, $g_9$, $g_{10}$, $g_{11}$, and $g_{12}$), five hybrid ($g_{13}$, $g_{14}$, $g_{15}$, $g_{16}$, and $g_{17}$) and ten composite ($g_{18}$, $g_{19}$, $g_{23}$, $g_{24}$, $g_{25}$, $g_{26}$, $g_{27}$, $g_{28}$, $g_{29}$, and $g_{30}$) test problems. For $g_6$, COGSA outperforms others except PSOGSA. While for $g_8$ and $g_{21}$, COGSA is the best algorithm except PTGSA. For $g_{20}$, GSA performs superior than others while COGSA is the better performer than PSOGSA and FVGGSA. For $g_{22}$, COGSA is better than GSA only. Regarding both mean and SD, COGSA outperforms other GSA variants for 19 test problems including four unimodal ($g_1$, $g_2$, $g_3$ and $g_4$), five multimodal ($g_5$, $g_9$, $g_{10}$, $g_{11}$, and $g_{12}$), three hybrid ($g_{13}$, $g_{14}$, and $g_{15}$) and seven composite ($g_{19}$, $g_{25}$, $g_{26}$, $g_{27}$, $g_{28}$, $g_{29}$, and $g_{30}$) problems. As per the third metric of the Table \ref{table:results_testbed2}, p-value of $137$ pairs are less than $0.01$. It indicates that there is a significant difference between these $137$ pairs. To investigate the exploitation ability of the COGSA over the complex problems of Testbed 2, convergence graphs are plotted in Figure \ref{fig:converge_cec} and Figure \ref{fig:converge_cec1}. These figures validate the fast convergence rate of the proposed COGSA over other considered GSA variants. The chaos embedded OBL strategy provides the most appropriate opposites through which COGSA gets the most promising regions very quickly. On the other hand,  chaotic gravitational constant improves the local search ability of COGSA through which it gets the quality solutions in the neighbourhood of the explored sub-optimal regions. The convergence graphs for the most difficult fitness landscapes of the composite functions ($g_{19}$, $g_{25}$, $g_{28}$, $g_{29}$, and $g_{30}$) validate its remarkable performance.

\newpage

\begin{center}
\begin{small}
\setlength{\LTleft}{-20cm plus -1fill}
\setlength{\LTright}{\LTleft}
\begin{longtable}{p{1.5cm}p{1.5cm}p{2.2cm}p{2.2cm}p{2.2cm}p{2.2cm}p{2.2cm}p{2.2cm}}
 \caption{Experimental results of considered algorithms over Testbed 2}
 \label{table:results_testbed2}
\endfirsthead
\caption* {\textbf{Table \ref{table:results_testbed2} Continued:}}
\endhead
\toprule[0.9pt]
\textbf{TP } & \textbf{metrics} &  \textbf{GSA}&  \textbf{PSOGSA}  &	\textbf{GGSA} &	\textbf{FVGGSA} & \textbf{PTGSA} &	\textbf{COGSA}  \\
 \midrule[0.9pt]

\multirow{3}{*} {$g_{1}$}			&	Mean	&	9.03E+05	&	2.32E+08	&	9.20E+05	&	4.95E+07	&	3.52E+05	&	\textbf{8.61E+04}	\\	&	SD	&	6.98E+05	&	2.82E+08	&	5.08E+05	&	9.34E+07	&	1.79E+05	&	\textbf{1.47E+05}\\ &	p-value	& $<2E-16$ & $<2E-16$ & $<2E-16$ & $<2E-16$ & $<2E-16$ &-\\[1ex]

\multirow{3}{*} {$g_{2}$}			&	Mean	&	7.68E+02	&	1.25E+09	&	8.76E+02	&	3.57E+09	&	4.28E+03	&	\textbf{6.39E+02}	\\		&	SD	&	9.65E+02	&	2.28E+09	&	1.01E+03	&	1.55E+09	&	6.89E+03	&	\textbf{8.61E+02}	\\ &	p-value	& $1$ & $<2E-16$ & $4.0E-05$ & $<2E-16$ & $6.8E-09$ &-\\[1ex]

\multirow{3}{*} {$g_{3}$}			&	Mean	&	5.31E+06	&	3.30E+08	&	5.97E+06	&	6.81E+07	&	2.35E+05	&	\textbf{7.95E+04}	\\	&	SD	&	6.74E+06	&	2.39E+08	&	7.43E+06	&	1.92E+07	&	8.42E+05	&	\textbf{3.56E+04}	\\ &	p-value	& $<2E-16$ & $<2E-16$ & $<2E-16$ & $<2E-16$ & $0.045$ &-\\[1ex]

\multirow{3}{*} {$g_{4}$}			&	Mean	&	1.22E+04	&	3.65E+04	&	1.20E+04	&	8.57E+04	&	4.74E+03	&	\textbf{4.17E+03}	\\	&	SD	&	6.53E+03	&	2.99E+04	&	6.07E+03	&	8.73E+03	&	1.71E+03	&	\textbf{1.23E+03}	\\ &	p-value	& $<2E-16$ & $<2E-16$ & $<2E-16$ & $<2E-16$ & $0.0034$ &-\\[1ex]

\multirow{3}{*} {$g_{5}$}			&	Mean	&	\textbf{2.00E+01}	&	\textbf{2.00E+01}	&	\textbf{2.00E+01}	&	\textbf{2.00E+01}	&	\textbf{2.00E+01}	&	\textbf{2.00E+01}	\\	&	SD	&	9.78E-05	&	1.26E-02	&	7.47E-05	&	8.46E-05	&	1.01E-02	&	\textbf{1.54E-05}	\\ &	p-value	& $<2E-16$ & $<2E-16$ & $<2E-16$ & $7.4E-05$ & $<2E-16$ &-\\[1ex]

\multirow{3}{*} {$g_{6}$}			&	Mean	&	2.12E+02	&	\textbf{1.77E+02}	&	2.12E+02	&	2.57E+02	&	2.12E+02	&	1.87E+02	\\	&	SD	&	2.44E+01	&	4.70E+01	&	2.44E+01	&	2.49E+01	&	2.39E+01	&	\textbf{2.06E+01}	\\ &	p-value	& $<2E-16$ & $0.0051$ & $1E-15$ & $<2E-16$ & $<2E-16$ &-\\[1ex]

\multirow{3}{*} {$g_{7}$}			&	Mean	&	3.78E+03	&	3.72E+03	&	3.77E+03	&	4.55E+03	&	3.84E+03	&	\textbf{3.55E+03}	\\		&	SD	&	\textbf{4.15E+02}	&	8.07E+02	&	4.85E+02	&	4.72E+02	&	4.47E+02	&	4.87E+02	\\ &	p-value	& $6.3E-07 $ & $2.0E-05$ & $0.014$ & $< 2E-16$ & $9.2E-14$ &-\\[1ex]

\multirow{3}{*} {$g_{8}$}			&	Mean	&	1.98E+02	&	1.37E+03	&	1.98E+02	&	4.15E+02	&	\textbf{1.01E+02}	&	1.15E+02	\\	&	SD	&	4.59E+01	&	1.00E+03	&	4.61E+01	&	7.28E+01	&	4.12E+01	&	\textbf{4.05E+01}	\\ &	p-value	& $<2E-16$ & $<2E-16$ & $<2E-16$ & $<2E-16$ & $4.9E-09$ &-\\[1ex]

\multirow{3}{*} {$g_{9}$}			&	Mean	&	2.00E+01	&	2.00E+01	&	2.00E+01	&	2.00E+01	&	2.00E+01	&	\textbf{2.00E+01}	\\	&	SD	&	5.54E-04	&	3.70E-05	&	5.49E-04	&	3.99E-04	&	5.18E-03	&	\textbf{1.48E-04}	\\ &	p-value	& $<2E-16$ & $<2E-16$ & $<2E-16$ & $1.9E-05$ & $< 2E-16$ &-\\[1ex]

\bottomrule[0.9pt]
\pagebreak
\toprule[0.9pt]
\textbf{TP } & \textbf{metrics} &  \textbf{GSA}&  \textbf{PSOGSA}  &	\textbf{GGSA} &	\textbf{FVGGSA} & \textbf{PTGSA} &	\textbf{COGSA}  \\
 \midrule[0.9pt]

\multirow{3}{*} {$g_{10}$}			&	Mean	&	1.82E+01	&	2.92E+01	&	1.80E+01	&	3.31E+01	&	1.77E+01	&	\textbf{1.62E+01}	\\	&	SD	&	2.02E+00	&	4.30E+00	&	2.17E+00	&	2.34E+00	&	2.56E+00	&	\textbf{2.02E+00}	\\ &	p-value	& $<2E-16$ & $<2E-16$ & $4.0E-16 $ & $1.9E-05$ & $ 6.5E-10$ &-\\[1ex]

\multirow{3}{*} {$g_{11}$}			&	Mean	&	4.18E-04	&	5.31E-01	&	5.24E-04	&	1.32E-02	&	1.16E-03	&	\textbf{2.45E-04}	\\	&	SD	&	4.95E-04	&	2.44E-01	&	7.24E-04	&	9.04E-03	&	1.09E-03	&	\textbf{2.46E-04}	\\ &	p-value	& $1$ & $<2E-16$ & $8.7E-08 $ & $<2E-16$ & $ <2E-16$ &-\\[1ex]

\multirow{3}{*} {$g_{12}$}			&	Mean	&	2.08E-01	&	1.09E+00	&	2.00E-01	&	1.67E+00	&	1.71E-01	&	\textbf{1.66E-01}	\\	&	SD	&	3.88E-02	&	9.93E-01	&	3.73E-02	&	7.97E-01	&	3.29E-02	&	\textbf{3.44E-02}	\\ &	p-value	& $<2E-16$ & $<2E-16$ & $<2E-16 $ & $<2E-16$ & $1$ &-\\[1ex]

\multirow{3}{*} {$g_{13}$}			&	Mean	&	1.22E+05	&	9.04E+06	&	1.09E+05	&	1.82E+06	&	2.36E+04	&	\textbf{1.91E+04}	\\	&	SD	&	5.80E+04	&	1.71E+07	&	5.39E+04	&	7.42E+05	&	2.46E+04	&	\textbf{8.45E+03}	\\ &	p-value	& $<2E-16$ & $<2E-16$ & $<2E-16$ & $<2E-16$ & $1$ &-\\[1ex]

\multirow{3}{*} {$g_{14}$}			&	Mean	&	1.53E+01	&	4.38E+01	&	1.44E+01	&	6.85E+01	&	8.61E+00	&	\textbf{8.00E+00}	\\	&	SD	&	8.28E+00	&	4.98E+01	&	5.62E+00	&	2.24E+01	&	1.89E+00	&	\textbf{1.32E+00}	\\ &	p-value	& $<2E-16$ & $<2E-16$ & $<2E-16$ & $<2E-16$ & $0.302$ &-\\[1ex]

\multirow{3}{*} {$g_{15}$}			&	Mean	&	2.24E+04	&	1.16E+06	&	2.20E+04	&	8.63E+04	&	1.52E+04	&	\textbf{1.30E+04}	\\		&	SD	&	8.20E+03	&	2.86E+06	&	9.01E+03	&	7.21E+04	&	3.77E+03	&	\textbf{2.50E+03}	\\ &	p-value	& $<2E-16$ & $<2E-16$ & $<2E-16$ & $<2E-16$ & $<2E-16$ &-\\[1ex]

\multirow{3}{*} {$g_{16}$}			&	Mean	&	2.98E+05	&	1.09E+07	&	5.21E+05	&	2.12E+06	&	9.43E+04	&	\textbf{6.67E+04}	\\	&	SD	&	1.50E+05	&	1.68E+07	&	1.18E+06	&	1.01E+06	&	\textbf{1.28E+05}	&	1.39E+05	\\ &	p-value	& $<2E-16$ & $<2E-16$ & $<2E-16 $ & $<2E-16$ & $7.2E-11$ &-\\[1ex]

\multirow{3}{*} {$g_{17}$}			&	Mean	&	4.98E+02	&	8.41E+06	&	5.24E+02	&	6.49E+02	&	4.54E+02	&	\textbf{3.41E+02}	\\	&	SD	&	3.69E+02	&	3.90E+07	&	4.36E+02	&	4.55E+02	&	\textbf{3.40E+02}	&	3.45E+02	\\ &	p-value	& $<2E-16$ & $<2E-16$ & $4.9E-15 $ & $<2E-16$ & $<2E-16$ &-\\[1ex]

\multirow{3}{*} {$g_{18}$}			&	Mean	&	1.51E+02	&	1.34E+02	&	1.49E+02	&	6.47E+02	&	1.50E+02	&	\textbf{1.26E+02}	\\		&	SD	&	1.20E+02	&	\textbf{5.48E+01}	&	1.15E+02	&	1.65E+02	&	1.10E+02	&	8.23E+01	\\ &	p-value	& $<2E-16$ & $<2E-16$ & $<2E-16$ & $<2E-16$ & $0.00099$ &-\\[1ex]

\bottomrule[0.9pt]	
\pagebreak	
	
\toprule[0.9pt]
\textbf{TP } & \textbf{metrics} &  \textbf{GSA}&  \textbf{PSOGSA}  &	\textbf{GGSA} &	\textbf{FVGGSA} & \textbf{PTGSA} &	\textbf{COGSA}  \\
 \midrule[0.9pt]

\multirow{3}{*} {$g_{19}$}			&	Mean	&	4.46E+05	&	7.61E+06	&	3.91E+05	&	2.40E+06	&	\textbf{2.99E+04}	&	\textbf{2.99E+04}	\\	&	SD	&	1.56E+05	&	1.39E+07	&	1.62E+05	&	8.00E+05	&	1.53E+04	&	\textbf{9.57E+03}	\\ &	p-value	& $<2E-16$ & $<2E-16$ & $<2E-16$ & $<2E-16$ & $0.012$ &-\\[1ex]

\multirow{3}{*} {$g_{20}$}			&	Mean	&	\textbf{3.25E+02}	&	1.15E+03	&	3.27E+02	&	4.87E+02	&	3.30E+02	&	3.50E+02	\\	&	SD	&	\textbf{1.02E+02}	&	3.55E+02	&	1.09E+02	&	2.96E+02	&	1.23E+02	&	1.56E+02	\\ &	p-value	& $<2E-16$ & $<2E-16$ & $<2E-16$ & $<2E-16$ & $1$ &-\\[1ex]

\multirow{3}{*} {$g_{21}$}			&	Mean	&	1.04E+02	&	1.43E+02	&	1.04E+02	&	2.36E+02	&	\textbf{1.02E+02}	&	1.03E+02	\\		&	SD	&	8.13E-01	&	2.54E+01	&	8.75E-01	&	2.48E+01	&	\textbf{5.95E-01}	&	6.34E-01	\\ &	p-value	& $<2E-16$ & $<2E-16$ & $<2E-16$ & $<2E-16$ & $7.2E-08$ &-\\[1ex]

\multirow{3}{*} {$g_{22}$}			&	Mean	&	5.11E+03	&	\textbf{3.02E+01}	&	4.94E+02	&	1.69E+03	&	1.23E+03	&	2.74E+03	\\	&	SD	&	3.98E+03	&	\textbf{2.56E+01}	&	4.90E+02	&	1.20E+03	&	1.46E+03	&	3.45E+03	\\ &	p-value	& $<2E-16$ & $<2E-16$ & $1$ & $<2E-16$ & $0.0167$ &-\\[1ex]

\multirow{3}{*} {$g_{23}$}			&	Mean	&	\textbf{1.00E+02}	&	6.18E+04	&	\textbf{1.00E+02}	&	4.08E+04	&	\textbf{1.00E+02}	&	\textbf{1.00E+02}	\\		&	SD	&	9.76E-08	&	1.26E+04	&	\textbf{2.24E-09}	&	6.59E+03	&	7.93E-02	&	1.22E-02	\\ &	p-value	& $<2E-16$ & $<2E-16$ & $<2E-16$ & $<2E-16$ & $<2E-16$ &-\\[1ex]

\multirow{3}{*} {$g_{24}$}			&	Mean	&	\textbf{1.00E+02}	&	5.71E+02	&	\textbf{1.00E+02}	&	1.72E+02	&	\textbf{1.00E+02}	&	\textbf{1.00E+02}	\\		&	SD	&	1.89E-10	&	1.43E+03	&	\textbf{5.27E-12}	&	1.77E+01	&	4.45E-04	&	2.11E-05	\\ &	p-value	& $<2E-16$ & $<2E-16$ & $<2E-16$ & $<2E-16$ & $<2E-16$ &-\\[1ex]

\multirow{3}{*} {$g_{25}$}			&	Mean	&	2.80E+02	&	3.67E+02	&	2.81E+02	&	2.41E+02	&	2.39E+02	&	\textbf{2.00E+02}	\\	&	SD	&	5.93E+01	&	3.66E+01	&	6.03E+01	&	7.13E+01	&	5.51E+01	&	\textbf{9.77E-04}	\\ &	p-value	& $0.00019$ & $<2E-16$ & $0.31418$ & $ 0.18677$ & $<2E-16$ &-\\[1ex]

\multirow{3}{*} {$g_{26}$}			&	Mean	&	2.00E+02	&	2.58E+02	&	2.00E+02	&	2.08E+02	&	2.00E+02	&	\textbf{2.00E+02}	\\	&	SD	&	1.12E-02	&	8.48E+00	&	9.95E-03	&	8.83E+00	&	1.17E-02	&	\textbf{2.50E-03}	\\ &	p-value	& $<2E-16$ & $<2E-16$ & $<2E-16$ & $ <2E-16$ & $<2E-16$ &-\\[1ex]

\multirow{3}{*} {$g_{27}$}			&	Mean	&	2.00E+02	&	1.46E+02	&	1.87E+02	&	1.92E+02	&	1.94E+02	&	\textbf{2.00E+02}	\\	&	SD	&	6.63E-03	&	4.92E+01	&	3.15E+01	&	2.38E+01	&	1.79E+01	&	\textbf{2.04E-08}	\\ &	p-value	& $<2E-16$ & $<2E-16$ & $<2E-16$ & $ <2E-16$ & $<2E-16$ &-\\[1ex]

\bottomrule[0.9pt]	
\pagebreak	
	
\toprule[0.9pt]
\textbf{TP } & \textbf{metrics} &  \textbf{GSA}&  \textbf{PSOGSA}  &	\textbf{GGSA} &	\textbf{FVGGSA} & \textbf{PTGSA} &	\textbf{COGSA}  \\
 \midrule[0.9pt]
	
\multirow{3}{*} {$g_{28}$}			&	Mean	&	2.62E+03	&	9.47E+02	&	1.47E+03	&	1.73E+03	&	1.80E+03	&	\textbf{2.00E+02}	\\	&	SD	&	1.06E+03	&	2.92E+02	&	4.17E+02	&	3.85E+02	&	4.27E+02	&	\textbf{7.15E-05}	\\ &	p-value	& $<2E-16$ & $<2E-16$ & $<2E-16$ & $ <2E-16$ & $<2E-16$ &-\\[1ex]

\multirow{3}{*} {$g_{29}$}			&	Mean	&	2.28E+03	&	4.31E+03	&	2.28E+03	&	2.42E+03	&	2.29E+03	&	\textbf{2.00E+02}	\\	&	SD	&	7.82E+02	&	1.05E+03	&	7.35E+02	&	7.62E+02	&	7.96E+02	&	\textbf{1.75E-04}	\\ &	p-value	& $<2E-16$ & $<2E-16$ & $<2E-16$ & $ <2E-16$ & $<2E-16$ &-\\[1ex]
	
\multirow{3}{*} {$g_{30}$}			&	Mean	&	1.05E+04	&	4.30E+05	&	1.16E+04	&	1.01E+05	&	3.04E+03	&	\textbf{1.77E+03}	\\	&	SD	&	7.36E+03	&	4.27E+05	&	1.02E+04	&	4.54E+04	&	5.90E+02	&	\textbf{1.78E+03}	\\ &	p-value	& $<2E-16$ & $<2E-16$ & $<2E-16$ & $ <2E-16$ & $1.5E-05$ &-\\[1ex]

\bottomrule[0.9pt]
\end{longtable}
 \end{small}
 \end{center}

\section{Conclusion}\label{sec:conclusion}
In this paper, chaos-embedded generalized opposition based learning is incorporated with the GSA framework to overcome its limitations regarding stagnation. Additionally, a dynamic gravitational constant supervised by the chaotic $\alpha$ parameter is introduced for a better trade-off between the exploration and exploitation abilities of GSA. In the initial phase of the search, chaos-based OBL provides a more robust exploration ability through which the algorithm gets the promising regions very quickly. On the other hand, the chaos attached gravitational constant recruits the optimal step sizes for the candidate solutions which significantly enhance the local search ability of the algorithm, especially for the middle and terminal phases of the search. The proposed COGSA is tested and validated over a wide range of benchmark problems including 23 classical benchmark problems as  Testbed $1$ and the set of CEC-2015 test suite along with 15 additional benchmark problems of CEC-2014 test suite as Testbed 2. The experimental and statistical analyses of the results confirm the supremacy of the proposed COGSA over the considered state-of-the-art algorithms and efficient GSA variants. The remarkable performance of COGSA on the highly complex composite problems encourages to study and develop more effective OBL strategies for a more robust and fast search mechanism in both continuous as well as discrete search space.

\end{document}